\documentclass[11pt]{article}
\usepackage{setspace}
\usepackage{graphicx,tikz,wrapfig,soul,color,epstopdf} 
\soulregister\cite7
\usepackage[a4paper, top=2.5cm,bottom=2.5cm,left=3cm,right=3cm,includeheadfoot,asymmetric]{geometry}
\usepackage{url,natbib}
\usepackage{times}
\usepackage[colorlinks=true,urlcolor=blue,linkcolor=red,citecolor=teal,backref=page]{hyperref}
\usepackage{booktabs,array,multirow}
\usepackage{tablefootnote,listings}
\usepackage{xcolor}
\usetikzlibrary{calc}
\usetikzlibrary{trees}
\usetikzlibrary{positioning}
\usepackage[edges]{forest}
\usetikzlibrary{arrows.meta}  
\usetikzlibrary{bayesnet,arrows}
\usetikzlibrary{positioning, automata}
\usetikzlibrary{shapes,fit,calc,positioning}
\tikzset{box/.style={draw, diamond, thick, text centered, minimum height=0.5cm, minimum width=1cm, text width=0.9cm}}
\tikzset{line/.style={draw, thick, -latex'}}

\usepackage[nohints]{minitoc}

\usepackage{array}
\usepackage{makecell}
\usepackage{float,xcolor,todonotes,tcolorbox}
\usepackage{mathtools}
\usepackage{amsmath, amssymb, natbib, graphicx, amsthm,amsfonts}
\usepackage{dsfont,enumitem}
\usepackage{footnote}
\usepackage{caption}
\usepackage{subcaption}
\usepackage{mwe}
\usepackage{cancel}

\makesavenoteenv{tabular}
\makesavenoteenv{table}
\newcommand{\alg}{\mathcal{A}}
\newcommand{\mech}{\mathcal{M}}
\newcommand{\real}{\mathbb{R}}
\newcommand{\bx}{\mathbf{x}}

\newcommand{\bX}{\mathbf{X}}

\newcommand{\CX}{\mathcal{X}}
\newcommand{\CY}{\mathcal{Y}}

\newcommand \expect {\mathop{\mbox{\ensuremath{\mathbb{E}}}}\nolimits}
\newcommand \prob {\mathop{\mbox{\ensuremath{\mathbb{P}}}}\nolimits}

\newcommand \traindata {\ensuremath{\mathbf{D}^T}}

\newtheorem{definition}{Definition}

\makeatletter
\newtheorem*{rep@theorem}{\rep@title}
\newcommand{\newreptheorem}[2]{%
	\newenvironment{rep#1}[1]{%
		\def\rep@title{\textbf{#2} \ref{##1}}%
		\begin{rep@theorem}}%
		{\end{rep@theorem}}}
\makeatother
\newreptheorem{theorem}{Theorem}
\newreptheorem{lemma}{Lemma}
\newreptheorem{proposition}{Proposition}
\newreptheorem{assumption}{Assumption}
\newreptheorem{corollary}{Corollary}

\newcommand{\sensitive}{\ensuremath{\mathbf{A}}}

\definecolor{existential}{rgb}{0.2,1,0.3}
\definecolor{random}{rgb}{0.9,0.9, 0.1}
\definecolor{affirmative}{rgb}{0.2,1,0.3}

\title{\textbf{The Fair Game:\\ Auditing \& Debiasing AI Algorithms Over Time}}
\author{Debabrota Basu and Udvas Das \thanks{The paper is published in 
    \textit{Cambridge Forum on AI: Law and Governance , Volume 1 , 2025 , p. e27}.
    DOI: \url{https://doi.org/10.1017/cfl.2025.8}}}
\date{Équipe Scool, Inria, Univ. Lille, CNRS, Centrale Lille, UMR 9189- CRIStAL\\
Lille, 59000 France}
\begin{document}

\maketitle


\begin{abstract}
An emerging field of AI, namely Fair Machine Learning (ML), aims to quantify different types of bias (also known as unfairness) exhibited in the predictions of ML algorithms, and to design new algorithms to mitigate them. Often, the definitions of bias used in the literature are observational, i.e. they use the input and output of a pre-trained algorithm to quantify a bias under concern. In reality,these definitions are often conflicting in nature and can only be deployed if either the ground truth is known or only in retrospect after deploying the algorithm. Thus,there is a gap between what we want Fair ML to achieve and what it does in a dynamic social environment. Hence, we propose an alternative dynamic mechanism,``Fair Game'',to assure fairness in the predictions of an ML algorithm and to adapt its predictions as the society interacts with the algorithm over time. “Fair Game” puts together an Auditor and a Debiasing algorithm in a loop around an ML algorithm. The ``Fair Game'' puts these two components in a loop by leveraging Reinforcement Learning (RL). RL algorithms interact with an environment to take decisions, which yields new observations (also known as data/feedback) from the environment and in turn, adapts future decisions. RL is already used in algorithms with pre-fixed long-term fairness goals. “Fair Game” provides a unique framework where the fairness goals can be adapted over time by only modifying the auditor and the different biases it quantifies. Thus,“Fair Game”aims to simulate the evolution of ethical and legal frameworks in the society by creating an auditor which sends feedback to a debiasing algorithm deployed around an ML system. This allows us to develop a flexible and adaptive-over-time framework to build Fair ML systems pre- and post-deployment.
\end{abstract}
\clearpage
\setcounter{parttocdepth}{4}
\doparttoc
\tableofcontents
\newpage

\section{Introduction}\label{sec:intro}

In today's era, learning machines are at the epicentre of technological, social, economic, and political developments, \textbf{{their continuous evaluation and alignment are critical concerns}}. 
In 2016, World Economic Forum\footnote{\url{https://www.weforum.org/meetings/world-economic-forum-annual-meeting-2016/}} recognised the study of learning machines, i.e. Machine Learning (ML), and its superset Artificial intelligence (AI) to be the driving force of the {fourth industrial revolution}\footnote{\url{https://www.weforum.org/agenda/2016/01/what-is-the-fourth-industrial-revolution}}. 
Reckoning of modern AI not only motivates development of efficient learning algorithms to solve real-life problems but also aspires socially aligned deployment of them. 
This aspiration has pioneered the theoretical and algorithmic developments leading to \textit{ethical}, \textit{fair}, \textit{robust}, and \textit{privacy-preserving AI}, in brief \textbf{responsible AI}~\citep{cheng2021socially,dpbook,fairMLbook,liu2021adversarial}. 
The frontiers of responsible AI are well-developed for static data distributions and models, but their extensions to dynamic environments are limited to Reinforcement Learning (RL) with stationary dynamics~\citep{sutton2018reinforcement}. 
Concurrently, the emerging trend of regulating AI poses novel regulations and quantifiers of risks induced by AI algorithms~\citep{gdpr,ccpa,hipaa,dsadma,madiega2021artificial}. Followed by deployment of GDPR\footnote{\url{https://gdpr-info.eu/}} in 2018 and upcoming EU AI act\footnote{\url{https://www.europarl.europa.eu/topics/en/article/20230601STO93804/eu-ai-act-first-regulation-on-artificial-intelligence}} in 2025-26, Europe pushes the frontiers of AI regulation, and propels the paradigm of {\textbf{algorithmic auditing}}\footnote{\url{https://auditing-fairness-tutorial.github.io/}}. Specially, EU AI act\footnote{\url{https://artificialintelligenceact.eu/}} discusses like any publicly used technology AI should undergo an audit mechanism, where we aim to understand the impacts and limitations of using this technology, and why they are caused. 
\textbf{But these two approaches are presently unwed.} 

Specifically, existing responsible AI algorithms fix a property (e.g. privacy, bias, robustness) first, then the theory is built to learn with these properties, and finally algorithms are designed to achieve optimal alignment~\citep{dpbook,liu2021adversarial,fairMLbook}. \textit{This present approach leaves little room for the auditor feedback to be incorporated in AI algorithms except the broad design choices.}

\paragraph{A Curious Case: \texttt{AIRecruiter}.} Let us consider an AI algorithm that uses a dataset of resumes and successful recruitments to learn whether an applicant is worth recruiting or not by an organisation. Multiple platforms, such as Zoho Applicant Tracking System\footnote{\url{https://www.zoho.com/fr/recruit/}}, LinkedIn job platform\footnote{\url{https://www.linkedin.com/business/talent/blog/applicant-tracking-system}}, are already used in practice. 
The designer would obviously want  the \texttt{AIRecruiter} algorithm to be accurate, regulation-friendly, and practically useful.
To be regulation-friendly, \texttt{AIRecruiter} has to consider the questions of social alignment, such as privacy, bias, and safety. 
In addition, the labour market and company's financial situations are dynamic. Thus, the recruitment policies and the drive to achieve social alignment of \texttt{AIRecruiter} evolves over time. 
\texttt{AIRecruiter} \textit{demonstrates one of the many practical applications, where social alignment of an AI algorithm over time becomes imperative.}

\paragraph{Unbiasedness (Fairness): Debiasing and Auditing.} If \texttt{AIRecruiter} is trained on historical data with a dominant demography, it would typically be
biased towards the `majority' and less generous to the minorities~\citep{barocas2023fairness}. This is a common problem in many applications: gender bias against women in Amazon hiring system\footnote{\url{https://heinonline.org/HOL/LandingPage?handle=hein.journals/cdozo44&div=7&id=&page=}}, economical bias against students from poorer background in SAT-score based college admission~\citep{kidder2001does}, racial bias against defendants of colour in the COMPAS crime recidivism prediction system~\citep{bagaric2019erasing}, to name a few. 
\textit{This issue invokes the question what is the fair or unbiased way of learning best sequence of decisions, and asks for conjoining ethics and AI.} 

\paragraph{Debiasing Algorithms.} Due to the ambiguous notions of fairness in society, researchers have proposed eclectic metrics for fairness (also called, unbiasedness) in offline and online settings~\citep{fairMLbook,buening2021meritocracy}.  Additionally, multiple algorithms are proposed to mitigate bias in predictions~\citep{hort2024bias}, which are mostly tailored for a given fairness metric. A few recent frameworks are proposed to analyse some group fairness metrics unitedly~\citep{chzhen2020fair,mangold2023differential,ghosh:hal-03770346} but \textit{it is not clear how to leverage them for a generic and dynamic bias mitigating algorithm design.}  

\paragraph{Auditing Algorithms.} The existing bias auditing algorithms follow two philosophies\footnote{A detailed exposition is available at \url{https://auditing-fairness-tutorial.github.io/}.}: \textit{verification} and \textit{estimation}. Verification-based auditors aim to check whether an algorithm achieves a bias level below a desired threshold while using as small number of the data as possible~\citep{goldwasser2021interactive,mutreja2023pac}. On the other hand, estimation-based auditors aim to directly estimate the bias level in the predictions of the algorithm while also being sample-efficient~\citep{bastani2019probabilistic,ghosh2021justicia,ghosh2022fvgm,yan2022active,ajarra2024active}. Though initial auditors used to evaluate the bias over full input data~\citep{galhotra2017fairness,bellamy2019ai}, distributional auditors have been developed to estimate bias over whole input data distribution~\citep{yan2022active,ajarra2024active}. 
Along with researchers in the community, we have developed multiple distributional auditors yielding Probably Approximately Correct (PAC) estimations of different bias metrics while looking into a small number of samples (Figure~\ref{fig:PAC_auditor}). Recently, ~\cite{ajarra2024active} have derived lower bounds on number of samples required to estimate different metrics to propose a Fourier transform based auditor that estimates all of the fairness metrics simultaneously. Authors also show that the auditor achieves constant manipulation proofness while scaling better than existing algorithms. This affirmatively concludes a quest for a universal and optimal bias auditors for offline algorithms\footnote{Another challenge for an auditor is achieving manipulation proofness~\citep{yan2022active,ajarra2024active}, which we are studying in \textbf{Regalia} project. }.

\paragraph{Need for a Fair Game: Adapting to Dynamics of Ethics and Society.} Now, if a multi-national organisation uses the \texttt{AIRecruiter} over the years, the nature of acquired data from new job applicants suffers distribution shift over time due to world economics, labour market, and other dynamics. Similarly, the regulations regarding bias also evolve from market to market, and time to time. For example, ensuring gender equality in recruitment has been well-argued since 1970s~\citep{arrow}, while ensuring demographic fairness through minority admissions is still under debate.

In present literature, auditors and debiasing algorithms are assumed to have static measures of bias oblivious to long-term dynamics, and to be non-interactive over time. 
\textit{As an application acquires new data over time, updates its model, and socially acceptable ethical norms and regulations also evolve, it motivates the vision of , and a continual alignment of AI with Auditor-to-Alignment loop.} 

In this context, first, we propose the \textbf{Fair Game} framework (Section~\ref{sec:framework}). Fair Game puts together an \textit{Auditor} and a \textit{Debiasing algorithm} in a loop around an ML algorithm, and treats the long-term fairness as a game of an auditor that estimates a bias report and a debiasing algorithm that uses this bias report to rectify itself further. 
We propose a Reinforcement Learning (RL) based approach to realise this framework in practice. 

Then, we specify a set of properties that Fair Game and its components should satisfy (Section~\ref{sec:properties}). 
\begin{enumerate}
    \item \textit{Data Frugal:} Data is the fuel of AI and often proprietary. Thus, it is hard to expect access to the complete dataset used by a large technology firm to audit its models and algorithms. Rather, often the regulations and auditing start by interacting with the software and understanding the bias induced by it. This follows access to a part of the dataset and models to audit it for rigorously. Thus, sample frugality is a fundamentally desired property of auditing algorithms. This become even more prominent in dynamic settings as the datasets, models and measures of bias change over time.
    \item \textit{Manipulation Proof:} As auditing is a time-consuming and often legal mechanism, one has to consider the opportunity to update models up to a certain extent in order to match the speed of fast changing AI landscape. Thus, an ideal auditor should be able to adapt to minor shifts or manipulation in the data distributions, model properties etc. So, we aim for building an auditor which satisfies constant manipulation proofness.
    \item \textit{Adaptive and Dynamic:} While designing an auditor for debiasing AI algorithm for real-life tasks, it should interact with the environment in a dynamic way, i.e. it should adapt to correct bias measures being in dynamic environment. For example, the notion of fairness can change over time, so an auditor should be dynamic to adapt to such properties that change over time. This is central conundrum to ensure long-term fairness. 
    \item \textit{Structured Feedback:} In existing auditors, we study statistically efficient auditors yielding accurate global estimates of privacy leakage, bias, and instability. All the ethical conundrums are not quantifiable from observed data, but are subjective. Thus, auditors might provide preferential feedback on different sub-cases of predictions~\citep{conitzerposition,Preference_collapse,safe_rlhf,yu2024rlhf}. 
    Additionally, an user who leverages an AI algorithm to make an informed decision, may want to the response aligned with his/her/their preferences. An auditor can help an AI algorithm to tune their responses with respect to these preference alignments. 
    \item \textit{Stable Equilibrium:} The other question is the existence of a stable equilibrium in Fair Game, which is a two-player game and both the players can manipulate the other. Under fixed bias measure but under dynamic shifts of data distribution and incremental model updates, Fair Game should be able to reach a stable equilibrium between the auditor and the AI algorithm. 
\end{enumerate}

Before proceeding to the contributions, in Section~\ref{sec:sota}, we elaborate the algorithmic background for algorithmic auditing of bias, debiasing algorithms, and their limitations as the basis of Fair Game. In Section~\ref{sec:rl}, we briefly introduce concepts of reinforcement learning as it is the algorithmic foundation of Fair Game. 
\section{Background: Static Auditing and Debiasing Algorithms}\label{sec:sota}
\begin{figure}[t!]
    \centering
    \includegraphics[width=0.8\textwidth,trim={0.5cm 3.5cm 1.2cm 4.2cm},clip]{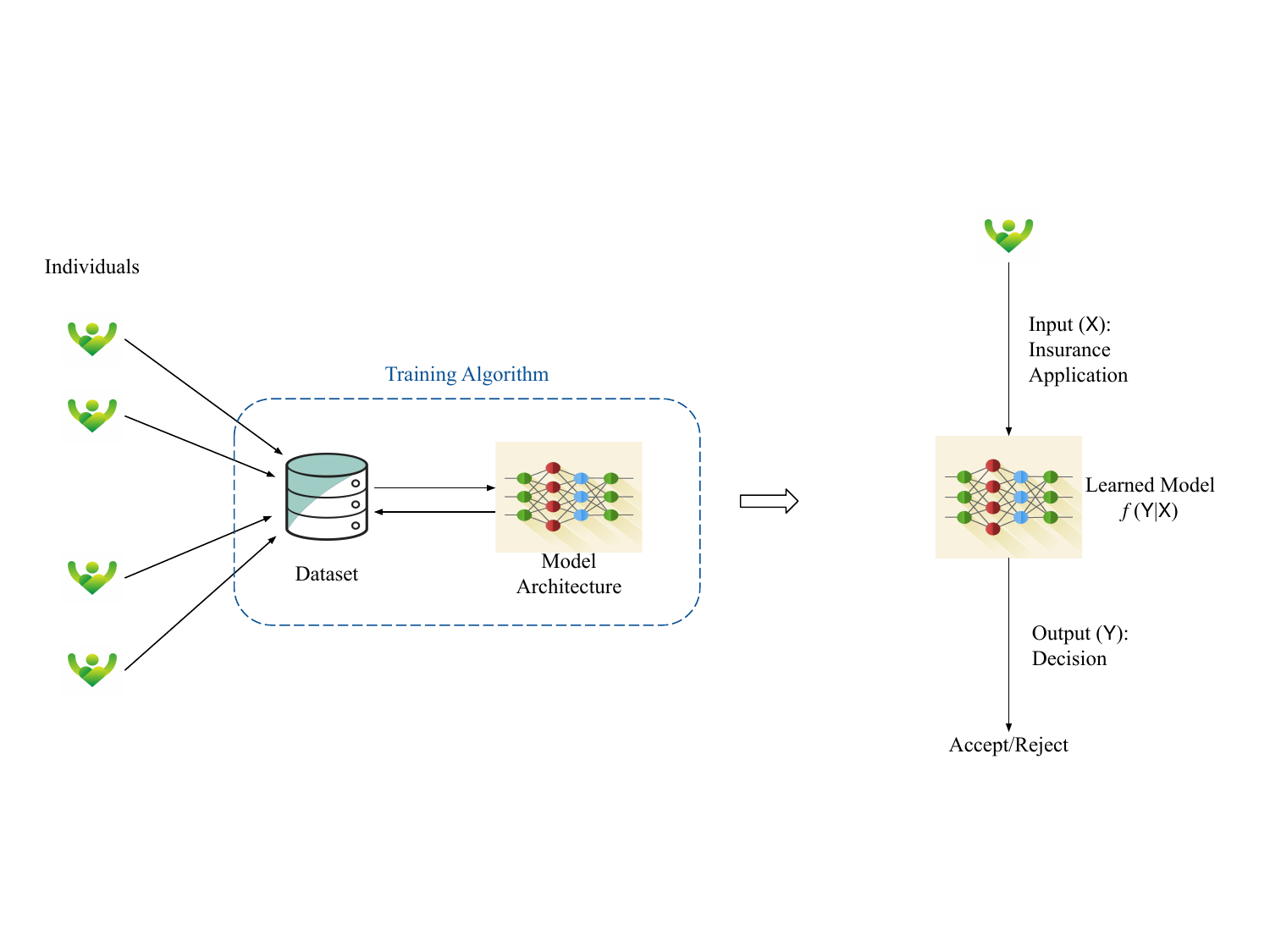}
    \caption{\texttt{AIRecruiter}: Training (left) and deploying (right) a machine learning model.}\label{fig:ml_model}
\end{figure}

\subsection{\textit{Learning to Predict from Data: Fundamentals of Machine Learning (ML) Models}}
An ML model~\citep{mohri2018foundations} is a function $f: \CX \rightarrow \CY$ that maps a set of input features $\bX \in \CX$ to an output $Y \in \CY$.\footnote{\textbf{Notations:} We denote sets/vectors by \textbf{bold} letters, and the distributions by \textit{calligraphic} letters. We express random variables in UPPERCASE, and an assignment of a random variable in lowercase.} For classifiers, the output space is a finite set of classes, i.e. $\{1,\ldots,k\}$. For regressors, the output space is a $d_o$-dimensional space of real numbers.  Training and deploying a Machine Learning (ML) model involves mainly \textit{four} components: (a) training dataset, (b) model architecture, (c) loss function, and (d) training algorithm. 

Commonly, an ML model $f$ is a parametric function, denoted as $f_{\theta}$, with parameters $\theta \in \real^d$, and is trained on a \textbf{training dataset} $\mathbf{D}^T$, i.e. a \textit{collection of $n$ input-output pairs $\{(\bx_i, y_i)\}_{i=1}^n$ generated from an underlying distribution} $\mathcal{D}$. The \textit{exact parametric form of the ML model} is dictated by the choice of \textbf{model architecture}, which includes a wide-variety of functions over last five decades. 
Training implies that given a model class $\mathcal{F}=\{f_\theta|\theta\in\Theta\}$, a loss function $l: \CY \times \CY \rightarrow \real_{\geq0}$, and training dataset $\traindata$, we aim to find the optimal parameter
\begin{align}\label{eq:erm}
    \theta^* \triangleq \arg\min_{\theta \in \Theta} \sum_{i=1}^n l(f_{\theta}(\bx_i), y_i).
\end{align}
A \textbf{loss function} \textit{measures badness of predictions made by the ML model with respect to the true output}. We commonly use cross-entropy, i.e. $l(f_{\theta}(\bx_i), y_i) \triangleq -y_i \log(f_{\theta}(\bx_i))$, as the loss function for classification. For regression, we often use the square loss $l(f_{\theta}(\bx_i), y_i) \triangleq \left( y_i - f_{\theta}(\bx_i)\right)^2$. Finally, \textit{an optimization algorithm is deployed to find the solution of the minimisation problem} in Equation~\eqref{eq:erm}. We refer to the optimiser as a \textbf{training algorithm}.

This procedure of training an ML model is called \textit{Empirical Risk Minimisation (ERM)}~\citep{vapnik1991principles, gyorfi2006distribution, devroye2013probabilistic, feldman2012agnostic}. ERM is at the core of successfully training decision trees to large deep neural networks, like LLMs. 
The key statistical concept behind using ERM to train parameteric ML models is that \textit{if we have used large enough training dataset and the parametric family (aka model architecture) is expressive enough, the trained model can predict accurately (with high probability) for unseen input points coming from the same or close enough data generating distributions.} 
This property is called \textit{generalisation ability} of an ML model and is often measured with its accuracy of predictions over a test dataset. A large part of statistical learning theory is dedicated to study this property for different types of data distributions, model architectures, and training algorithms~\citep{mohri2018foundations,dperm,dpadmm,dpregression}. 

In Figure~\ref{fig:ml_model}, we provide a schematic of this training and deployment schematic of ML models in the context of \texttt{AIRecruiter}. Specifically, \texttt{AIRecruiter} uses a historical dataset of resumes and future performance of job seekers to train a recruitment predicting ML model (\textit{left} side of Figure~\ref{fig:ml_model}). After successful training of the ML model, when it is deployed in practice, a resume of a candidate is sent through it and the model recommends accepting or rejecting the candidate (\textit{right} side of Figure~\ref{fig:ml_model}).  This is well-known as the binary classification problem. For example, \cite{buening:hal-03445971} showed similar mechanism and its nuances of gender and demographic bias in the context of college admissions. They use 15 years of data from Norwegian college admissions and examination performances to show that the historical data and fairness-oblivious ML models trained on it exhibit different types of bias under testing. We are aware of multiple such examples world-wide, such as racial bias in crime recidivism prediction in the COMPAS case~\citep{compas}, gender bias in translating and completing phrases involving occupations by LLMs~\citep{gorti2024unboxing}, economic bias in  SAT-score based college admissions~\citep{kidder2001does} to name a few.  

But discriminations in all these cases are prohibited up to different extents by laws of different countries~\citep{madiega2021artificial, veale2021demystifying, fiss1970theory, blumrosen1967duty} and also are ethically unfair, in general~\citep{novelli2023taking}. This calls for design of bias auditing and debiasing ML algorithms.

\subsection{\textit{Debiasing Algorithms: State-of-the-art}}\label{sec:debias}
A \textbf{debiasing algorithm} observes the input and output of an ML algorithm, the different quantifiers of bias estimated by the auditor, and (if possible) the architecture of the ML algorithm to recalibrate the predictions of the ML algorithm such that the different quantifiers of bias are minimised~\citep{lohia2019bias,chouldechova2020snapshot,mehrabi2021survey,barocas2023fairness,caton2024fairness}.

\paragraph{Measures of Bias.}  Before proceeding to the debiasing algorithms, we provide a brief but formal introduction to different measures of bias. 

In order to explain measures of bias, we consider a binary classification task (e.g. \texttt{AIRecruiter}) on a dataset $\traindata$ as a collection of triples $(\mathbf{X}, \mathbf{A}, Y)$ generated from an underlying distribution $\mathcal{D}$. $\mathbf{X} \triangleq\left\{X_1, \ldots, X_{m_1}\right\}$ are non-sensitive features whereas $\mathbf{A} \triangleq\left\{A_1, \ldots, A_{m_2}\right\}$ are categorical sensitive features. $Y \in\{0,1\}$ is the binary label (or class) of ($\mathbf{X}, \mathbf{A})$. Each non-sensitive feature $X_i$ is sampled from a continuous probability distribution $\mathcal{X}_i$, and each sensitive feature $A_j \in\left\{0, \ldots, N_j\right\}$ is sampled from a discrete probability distribution $\mathcal{A}_j$. We use $(\mathbf{x}, \mathbf{a})$ to denote the feature-values of $(\mathbf{X}, \mathbf{A})$. For sensitive features, a valuation vector $\mathbf{a}=\left[a_1, . ., a_{m_2}\right]$ is called a compound sensitive group. For example, consider $\mathbf{A}=\{$race, sex$\}$ where race $\in\{$Asian, Color, White$\}$ and sex $\in$ $\{$female, male$\}$. Thus $\mathbf{a}=$ [Asian, female] is a compound sensitive group. We represent a binary classifier trained on the dataset $\mathbf{D}$ as $f:(\mathbf{X}, \mathbf{A}) \rightarrow \hat{Y}$. Here, $\hat{Y} \in\{0,1\}$ is the predicted class of $(\mathbf{X}, \mathbf{A})$.

\noindent\textit{I. Measures of Independence.} \textbf{The prediction $\widehat{Y}$ of a classifier for an individual is independent of its sensitive feature $A$.} 
Mathematically, if $\widehat{Y}$ is binary variable, independence implies that for all $\mathbf{a}, \mathbf{b}$, $\Pr[\widehat{Y}=1|\sensitive=\mathbf{a}] =  \Pr[\widehat{Y}=1|\sensitive=\mathbf{b}]$. 
Statistical/demographic parity~\citep{corbett2017algorithmic, feldman2015certifying, kamishima2012fairness, zemel2013learning} measures deviation from independence of a classifier for a given data distribution. 
\begin{definition}[Statistical Parity]
$\mathrm{SP} \triangleq \max_{\mathbf{a}} \Pr[\widehat{Y}=1|\sensitive=\mathbf{a}] - \min_{\mathbf{a}}\Pr[\widehat{Y}=1|\sensitive=\mathbf{a}]$.
\end{definition}
\begin{definition}[Demographic Parity]
$\mathrm{DP} \triangleq \frac{\min_{\mathbf{a}}\Pr[\widehat{Y}=1|\sensitive=\mathbf{a}]}{\max_{\mathbf{a}}\Pr[\widehat{Y}=1|\sensitive=\mathbf{a}]}$.
\end{definition}

\noindent The use of aforementioned metrics ensures equality of outcome across demographics. Though they can lead to accepting random people from majority and qualified people from minority, due to sample size disparity.

\noindent\textit{II. Measures of Sufficiency.} \textbf{The probability of positive outcome $\widehat{Y}$ for an individual given the true outcome is positive $Y=1$ should be independent of its sensitive feature $A$.} 
Mathematically, if $\widehat{Y}$ and $Y$ are binary variables, separation (equality of opportunity or equalised odds) implies that for all $\mathbf{a}, \mathbf{b}$, $\Pr[\widehat{Y}=1|Y=1, \sensitive=\mathbf{a}] =  \Pr[\widehat{Y}=1|Y=1, \sensitive=\mathbf{b}]$. 
Separation metrics measure deviation from conditional independence of a classifier for a data distribution. 
\begin{definition}[Equalised odds]
$\mathrm{EO} \triangleq \max_{a,b} \lvert \Pr[\widehat{Y}=1| Y=1, \sensitive=\mathbf{a}] - \Pr[\widehat{Y}=1| Y=1, \sensitive=\mathbf{b}] \rvert$.
\end{definition}

\noindent Incorporating EO~\citep{hardt2016equality, pleiss2017fairness} as a measure of bias ensures equality of outcome for eligible individuals across demographics. But on the other hand, the true outcome $Y$ is often not known in reality.

\noindent\textit{III. Measures of Calibration.} \textbf{A classifier prediction $\widehat{Y}$ should be calibrated such that conditional probability of the true outcome $Y=0/1$ should be independent of its sensitive feature $A$ given the prediction being $\widehat{Y}=0/1$.}
Mathematically, if $\widehat{Y}$ and $Y$ are binary variables, separation implies that for all $a, b$, $\Pr[Y=1|\widehat{Y}=1, \sensitive=\mathbf{a}] =  \Pr[Y=1|\widehat{Y}=1, \sensitive=\mathbf{b}]$ and $\Pr[Y=1|\widehat{Y}=0, \sensitive=\mathbf{a}] =  \Pr[Y=1|\widehat{Y}=0, \sensitive=\mathbf{b}]$. 
\begin{definition}[Predictive Value Parity (PVP)]
    \begin{align*}
        \mathrm{PVP} \triangleq \max \{ &\max_{\mathbf{a}}  \Pr[Y=1|\widehat{Y}=1, \sensitive=\mathbf{a}] - \min_{\mathbf{a}}  \Pr[Y=1|\widehat{Y}=1, \sensitive=\mathbf{a}],  \\
                &\max_{\mathbf{a}}  \Pr[Y=1|\widehat{Y}=0, \sensitive=\mathbf{a}] - \min_{\mathbf{a}}  \Pr[Y=1|\widehat{Y}=0, \sensitive=\mathbf{a}]
        \}
    \end{align*}
\end{definition}

\noindent Such calibration measures like PVP~\citep{chouldechova2017fair,hardt2016equality} equalizes chance of success given acceptance. But the acceptance largely depends on the choice of the classifier's utility function, which can be bias inducing.

There are other causal measures of bias than these three families of observational fairness metrics. We refer to \citep{chouldechova2017fair,mehrabi2021survey,barocas2023fairness} for further details on them.

\paragraph{Debiasing Algorithms.} Given a measure of bias, there are three types of debiasing algorithms proposed in the literature: (a) pre-processing, (b) in-processing, and (c) post-processing. These three families of algorithms intervene at three different parts of an ML model, i.e. training dataset, loss/training algorithm, and final predictions post-deployment. 


\noindent\textit{I. Pre-processing algorithms.} These algorithms recognise that often the bias is induced from the historically biased data used for training the algorithm~\citep{barocas2023fairness, kidder2001does, bagaric2019erasing}. When the data includes a lot of samples for a demographic majority and very little for other minorities, then under ERM framework that tries to minimise the average loss to find the best parameters often lead to learning the patterns accurately for the majority and ignoring that of the minorities. Thus, pre-processing algorithms try to transform the training dataset and create a `repaired' dataset~\citep{salimi2019data,salimi2019capuchin,salimi2019interventional,feldman2015certifying,gordaliza2019obtaining, hajian2012methodology, heidari2018preventing, kamiran2012data, luong2011k}. The advantage of pre-processing algorithms is that once the dataset is repaired, any model architecture and training algorithm can be used on top of it. The disadvantage is that it requires accessing and refurbishing the whole input dataset, which might include millions and billions of samples for large-scale deep neural networks. Thus, it becomes computationally intensive and requires retraining the downstream model for any update in bias measure. 

\noindent\textit{II. In-processing algorithms.} These algorithms aim to repair the fact that the classical ERM \textit{tries to accurately learn on an average over the whole data distribution.} Thus, they either reweigh the input samples according to their sensitive features~\citep{calders2013unbiased, jiang2020identifying, kamiran2012data}, or modify the loss to maximise both fairness and accuracy~\citep{agarwal2018reductions, celis2019classification, chierichetti2019matroids, cotter2019optimization}. Some of the recent works have try and design optimisation algorithms that consider the fairness and accuracy simultaneously and iteratively during training. The advantage of these algorithms is that they achieve tightest fairness-accuracy trade-offs among the three families of debiasing algorithms. The disadvantage is that they require retraining an already established ML model from scratch, which is often time consuming, economically expensive, and hard to convince the companies for whom models are central products.

\noindent\textit{III. Post-processing algorithms.} The third family of debiasing algorithms approach the problem post-training an ML model~\citep{chouldechova2017fair, hebert2018multicalibration, kim2018fairness, kleinberg2016inherent, liu2019implicit, liu2017calibratedfairnessbandits}. They recognise the impact of bias of an ML model can be stopped if only the predictions can be recalibrated according to their sensitive features and corresponding input-output distributions.  Thus, post-processing approaches often apply transformations to model's output to improve fairness in predictions. If a debiasing algorithm can treat the ML model without accessing the data or training procedure, this is the only feasible approach to debias the ML model. Thus, it is the most flexible family of methods and can be used as a wrapper around existing algorithms. The only disadvantage is that we know it is not possible in one-shot to debias ML models with post-processing methods, and might lead to sub-optimal accuracy levels in some cases. 

For more details on the debiasing algorithms, we refer interested readers to detailed surveys and books published on this topic over years~\citep{chouldechova2020snapshot,mehrabi2021survey,barocas2023fairness,caton2024fairness}.

\paragraph{Limitations.} We do not have a framework to accommodate the bias auditor feedback to improve debiasing of the algorithm under audit. This is fundamental to bridge the regulation-based and learning-based approaches to debiasing of an ML algorithm. 
 
\subsection{\textit{Auditors of Bias: State-of-the-Art}}
An \textbf{auditor} looks into the input and output pairs of an ML system and tries to measure different types of bias. Any publicly used technology presently undergoes an audit mechanism, where we aim to understand the impacts and limitations of using that technology, and why they are caused. In last decade, this has slowly but increasingly motivated development of statistically efficient auditors of bias, risk, and privacy leakage caused by an ML algorithm. 
Though the initial auditors were specific to a training dataset and used the whole dataset to compute an estimate of bias~\citep{aif360-oct-2018,pentyala2022privfairfl}, a need of sample-efficient estimation of bias has been felt. This has led to a Probably Approximately Correct (PAC) auditor that uses a fraction of data to produce the estimates of bias which are correct for the whole input-output data distributions~\citep{albarghouthi2017fairsquare,bastani2019probabilistic,ghosh2021justicia,yan2022active,ajarra2024active}. 
Here, we formally define a Probably Approximately Correct (PAC) auditor of a distributional property of an ML model, such as different bias measures. 
\begin{definition}[PAC auditor~\citep{ajarra2024active}]\label{def:pac_auditor}
    Let $\mu: \mathbf{D}^T \times f_{\theta} \rightarrow \real$ be a computable\footnote{Any distributional property of an ML model, including risk~\citep{paltrinieri2019learning,assaf2020utilization,galindo2000credit}, individual fairness~\citep{pessach2022review,chouldechova2018frontiers}, and group fairness~\citep{chouldechova2020snapshot}, is computable given the existence of the mean estimators with uniformly random samples.} distributional property of an ML model $f_{\theta}$. An algorithm $\mathcal{A}$ is a \textit{PAC auditor} of property $\mu$ if for any $\epsilon, \delta \in (0,1)$, there exists a function $m(\epsilon, \delta)$ such that $\forall m \geq m(\epsilon, \delta)$ samples drawn from $\mathcal{D}$, it outputs an estimate $\hat{\mu}_m$ satisfying 
    \begin{align}
        \prob(|\hat{\mu}_m - \mu| \leq \epsilon) \geq 1- \delta\,.
    \end{align}
\end{definition}

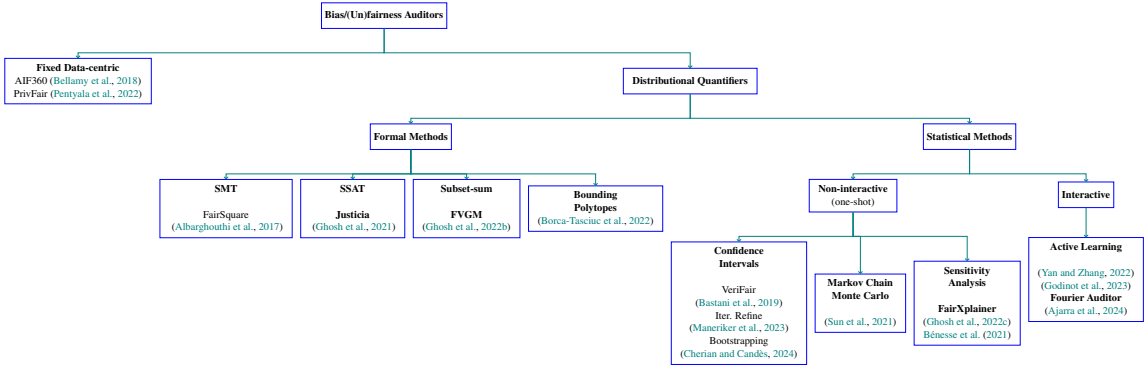
\begin{figure}[t!]
\centering
\resizebox{!}{0.32\textwidth}{
\begin{forest}
for tree={
draw=blue,      
fill=none,   
       minimum height = 5.5ex,
        minimum width = 2em,
edge={semithick, -{Straight Barb[scale=0.8]},draw=teal},    
forked edge,        
    l sep = 11mm,    
    s sep = 3mm,    
    where level = 4{s sep=1mm}{},
 fork sep = 9mm,    
           }
[\textbf{Bias/(Un)fairness Auditors}
    [\begin{tabular}{c}\textbf{Fixed Data-centric}\\ {AIF360~\citep{aif360-oct-2018}}\\  {PrivFair~\citep{pentyala2022privfairfl}}\end{tabular}]
    [\begin{tabular}{c}\textbf{Distributional Quantifiers}\end{tabular},fit=band
       [\textbf{Formal Methods}
            [\begin{tabular}{c}\textbf{SMT}\\ \\  FairSquare\\~\citep{albarghouthi2017fairsquare}\end{tabular}]
                [\begin{tabular}{c}\textbf{SSAT}\\~\\  \textbf{Justicia}\\~\citep{ghosh2021justicia}\end{tabular}]
                [\begin{tabular}{c}\textbf{Subset-sum}\\~\\ \textbf{FVGM}\\ \citep{ghosh2022algorithmic}\end{tabular}]
                [\begin{tabular}{c}\textbf{Bounding}\\ \textbf{Polytopes}\\  \citep{borca2022provable}\end{tabular}]]
        [\textbf{Statistical Methods},fit=band
            [\begin{tabular}{c}\textbf{Non-interactive}\\ (one-shot)\end{tabular}
                [\begin{tabular}{c}\textbf{Confidence}\\\textbf{Intervals}\\~\\  VeriFair\\ \citep{bastani2019probabilistic}\\ Iter. Refine \\\citep{maneriker2023online}\\ Bootstrapping\\\citep{cherian2023statistical}\end{tabular}]
                [\begin{tabular}{c}\textbf{Markov Chain}\\\textbf{Monte Carlo}\\~\\ {\citep{sun2021probabilistic}}\end{tabular}]
                [\begin{tabular}{c}\textbf{Sensitivity}\\\textbf{Analysis}\\~\\ 
                \textbf{FairXplainer}\\\citep{Ghosh_23}\\
              {\cite{benesse2021}}\end{tabular}]
            ]
            [\textbf{Interactive}
                [\begin{tabular}{c}\textbf{Active Learning}\\~\\  \citep{yan2022active}\\\citep{godinot}\\
                 \textbf{Fourier Auditor}\\\citep{ajarra2024active}\end{tabular}]
            ]
        ]
    ]
]
\end{forest}}
    \caption{A taxonomy of bias auditors.}\label{fig:sota_auditors}
\end{figure}

In Figure~\ref{fig:sota_auditors}, we provide a brief taxonomy of the existing bias auditing algorithms. 

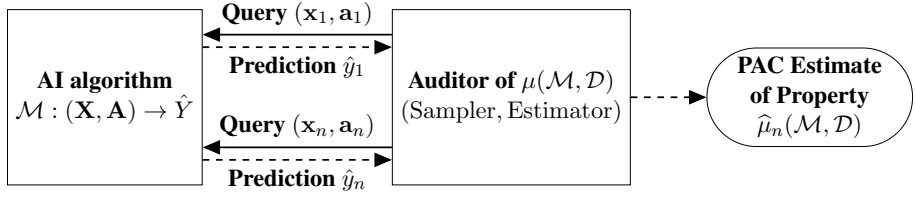
\begin{figure}[t!]
	\centering
	\resizebox{0.8\textwidth}{!}{
	\begin{tikzpicture}[
		node distance=5mm and 30mm,
		box/.style = {draw, minimum height=28mm, align=center},
		sy+/.style = {yshift= 2mm}, 
		sy-/.style = {yshift=-2mm},
		every edge quotes/.style = {align=center}
		]
		\node (n1) [box]             {\textbf{AI algorithm}\\$\mathcal{M}: (\mathbf{X}, \mathbf{A}) \rightarrow \hat{Y}$};                        
		\node (n2) [box,right=of n1] {\textbf{Auditor of} $\mu(\mathcal{M},\mathcal{D})$\\$(\mathrm{Sampler}, \mathrm{Estimator})$};
		\node (n3) [right=of n2, xshift=-18mm, draw, rounded rectangle, align=center] {\textbf{PAC Estimate}\\\textbf{of Property}\\ $\widehat{\mu}_n (\mathcal{M},\mathcal{D})$};
		\draw[thick,->, dashed]  ([yshift=8mm] n1.east) -- ([yshift=8mm] n2.west) node [midway, below] {\textbf{Prediction} $\hat{y}_1$};
		\draw[thick,->]  ([yshift=10mm] n2.west) -- ([yshift=10mm] n1.east) node [midway, above] {\textbf{Query} $(\mathbf{x}_1, \mathbf{a}_1)$};
		\draw[thick,->, dashed]  ([yshift=-10mm] n1.east) -- ([yshift=-10mm] n2.west) node [midway, below] {\textbf{Prediction} $\hat{y}_n$};
		\draw[thick,->]  ([yshift=-8mm] n2.west) -- ([yshift=-8mm] n1.east) node [midway, above] {\textbf{Query} $(\mathbf{x}_n, \mathbf{a}_n)$};
		\draw[thick,->, dashed]  (n2.east) -- (n3.west);
	\end{tikzpicture}}
	\caption{A generic schematic of a PAC auditor of a distributional property $\mu$.}\label{fig:PAC_auditor}
\end{figure}

\paragraph{Components of a PAC Auditor.} As illustrated in Figure~\ref{fig:PAC_auditor}, any PAC auditor consists of two components: (a) \textit{sampler} and (b) \textit{estimator}. 

The \textbf{sampler} selects a bunch of input-output pairs from a dataset, which might or might not be same with the training dataset, and then send them further to query the ML model under audit. 
The information obtained by querying the algorithm depends on the access of the auditor. 
For example, an internal auditor of an organisation can obtain much detailed information, such as confidence of predictions, gradients of loss at the query points etc. This is called a white-box setting. 
On the other hand, an external auditor (e.g. a public body or a third-party company) might get only the predictions of the ML model for the queried samples. This is called a black-box setting. 
Initially, the auditing algorithms used only uniformly random samples from a dataset (e.g. the `Formal Methods' and `Non-interactive' auditors in Figure~\ref{fig:sota_auditors}). But we know from statistics and learning theory that uniformly random sampling is less sample efficient than active sampling methods for most of the estimation problems. 
Specifically, the minimum number of samples required to PAC estimate mean of a distribution under uniform sampling is $\Omega(\epsilon^{-2}\ln(1/\delta))$, whereas the same for active sampling is $\Omega(\epsilon^{-1}\ln(1/\delta))$~\citep{yan2022active}. Thus, to obtain an error of $1\%$ or below in PAC mean estimation, we get $100$X decrease in the required number of samples. 
This has motivated design of active sampling mechanisms for auditing bias~\citep{yan2022active,godinot,ajarra2024active}, and even other properties, like stability under different perturbations~\citep{ajarra2024active}.

The \textbf{estimator} is the other fundamental component of an auditor. 
Typically, estimators try and quantify a specific measure of bias~\citep{albarghouthi2017fairsquare,ghosh2023biased} or a family of bias measures~\citep{bastani2019probabilistic,ghosh2021justicia,ghosh2022algorithmic,ajarra2024active}. 
All of these estimators use the queried samples and their corresponding outputs to compute the bias measures. 
Designing efficient and stable estimators require understanding the structural properties of different bias measures properly and leveraging it in the algorithmic scheme.  
For example, \cite{bastani2019probabilistic} address bias estimation of programs as an SMT problem, \cite{ghosh2021justicia} treats the same for any classifier as Stochastic SAT (SSAT) problem, whereas \cite{bastani2019probabilistic,yan2022active} brings it back to a set of conditional mean estimation problems. 
\cite{ghosh2023biased} and \cite{benesse2021} additionally aim for estimation of bias with attribution to features using global sensitivity analysis techniques from functional analysis. 
\cite{ajarra2024active} generalise this further by observing all the metrics of stability and bias are in the end impacts of different perturbations to the model distributions and can be computed using Fourier transformation of the input-output distribution of an ML model under audit. 

\paragraph{Limitations.} The present bias auditors, even the sample-efficiency wise optimal ones~\citep{ajarra2024active}, can only audit static/offline AI algorithms accurately. The question to extend them for dynamic algorithms is still open. This is critical to create an auditor-to-alignment loop to audit and debias ML models over time.

\subsection{\textit{Learning with Feedback: A Primer on Reinforcement Learning (RL)}}\label{sec:rl}
As we want to create a feedback mechanism between the auditor and debiasing algorithm, and use their feedback to align ML models over time, we need to study the Reinforcement Learning (RL) paradigm of ML that aims to learn about a dynamic environment while using only iterative feedback from the environment~\citep{altman1999constrained,sutton2018reinforcement}. This ability of RL has been recognised and thus, it has been studied for long-term fairness and sequential decision making problems, such as college admissions over years~\citep{buening2021meritocracy}. The advantage is that `fair' RL algorithms allow us to refine biased decisions over time but like all existing debiasing algorithms they often need a fixed measure of bias and are tailor-made to optimise it~\citep{gajane2022survey}.

In RL~\citep{sutton2018reinforcement}, a learning \textit{agent} sequentially interacts with an \textit{environment} by taking a sequence of \textit{actions}, and subsequently observing a sequence of \textit{rewards} and changes in her \textit{states}. 
Her goal is to compute a sequence of actions that yields as much reward as possible given a time limit. In other words, the agent aims to discover an optimal \textit{policy}, i.e. an optimal mapping between her state and corresponding feasible actions leading to maximal accumulation of rewards.
Two principal formulations of RL are Bandits~\citep{lattimore2020bandit} and Markov Decision Processes (MDPs)~\citep{altman1999constrained}.

\textbf{Bandit} is an archetypal setting of RL with one state and a set of actions~\citep{lattimore2020bandit}. Each action corresponds to an unknown reward distribution. \textit{The goal of the agent is to take a sequence of actions that both discovers the optimal action and also allows maximal accumulation of reward.} The loss in accumulated rewards due to the unknown optimal action is called \textit{regret}. Bandit algorithms are commonly designed with a theoretical analysis yielding an upper bound, i.e. a limit on the worst-case regret that it can incur~\citep{azize:hal-03781600,azize:hal-04215474,azize:hal-04611650,basu2019differential,basu2022bandits}. From information theory and statistics, we also know that this regret cannot be minimised more than a certain extent. This is called the \textit{lower bound} on regret and indicates the fundamental hardness of the bandit. An algorithm is called \textbf{optimal} if its regret upper bound matches the lower bound up to constants.

In addition to Bandits, \textbf{MDPs} include multiple states and a transition dynamics that dictates how taking an action transits the agent from one state to the other~\citep{altman1999constrained}. An added challenge is to learn the transition dynamics, and optimise the future rewards with it. The latter is called the \textit{planning} problem. Thus, it requires generalising the optimal algorithm design tricks for bandits to handle the unknown transition dynamics, and also to use an efficient optimiser to solve the planning problem. \textit{Since RL considers the effects of sequential observations and learning with partial feedback from a dynamic environment, it serves as the perfect paradigm to investigate the auditor-to-alignment over time.}

\begin{figure}[t!]
	\centering
	\includegraphics[width=0.6\linewidth,trim={5cm 11.5cm 2cm 12.5cm},clip]{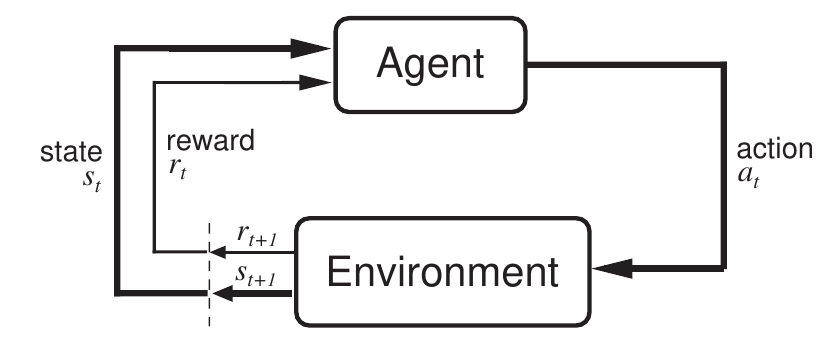}\vspace*{-.5em}
	\caption{The feedback loop in Reinforcement Learning (RL).}\vspace*{-1em}
\end{figure}

\textit{We specifically treat the Fair Game framework} (elaborated in Section~\ref{sec:framework}) as \textbf{performing RL in stochastic games.} On a positive note, success of many practical RL systems emerges in multi-agent settings, including playing games such as chess and Go~\citep{silver2016mastering,silver2017mastering}, robotic manipulation with multiple connected arms~\citep{gu2017deep}, autonomous vehicle control in dynamic traffic and automated production facilities~\citep{yang2020multi,eriksson:hal-03150823,eriksson:hal-03770369}. 
Further advances in these problems critically depend on developing stable and agent incentive-compatible learning dynamics in multi-agent environment. 
Unfortunately, the mathematical framework upon which classical RL depends on is inadequate for multiagent learning, since it assumes an agent’s environment is stationary and does not contain any adaptive agents. 
Classically, in multi-agent RL, these systems are treated as a stochastic game with a shared utility~\citep{brown1951iterative,shapley1953stochastic}. 
As an adaptive agent in this game only observes the outcomes of other's actions, the sequential and partial feedback emerges naturally. 
This connection has led to a growing line of works to understand limits of designing provably optimal RL algorithms for stochastic games~\citep{liu2022sample,giannou2022convergence,daskalakis2023complexity}. 
The existing analysis are often RL algorithm specific, i.e. they assume all the agents have agreed to play the same algorithm~\citep{giannou2022convergence}. 
On the other hand, the lower bounds quantifying the statistical complexity of these problems are mostly available for either zero-sum games~\citep{zhang2020model,fiegel2023adapting}, or large number of players (called mean-field games)~\citep{elie2020convergence}. 

\section{Fair Game Framework:  Auditing \& Debiasing Over Time}\label{sec:framework}
Now, we formulate the Fair Game framework that aims to resolve two issues:
\begin{enumerate}[leftmargin=*]
    \item incorporating the auditor feedback in the debiasing algorithm of an ML model,
    \item adapting to dynamics of society and ethical norm, and iteratively resolving the impacts of deploying ML models over time.
\end{enumerate}
First, we provide a high-level overview of the framework and its components. 
Then, we further formulate the framework and the corresponding problem statement rigorously. 

\begin{figure}[t!]
	\centering
	\includegraphics[width=0.8\linewidth,trim={0cm 1.5cm 0 0cm},clip]{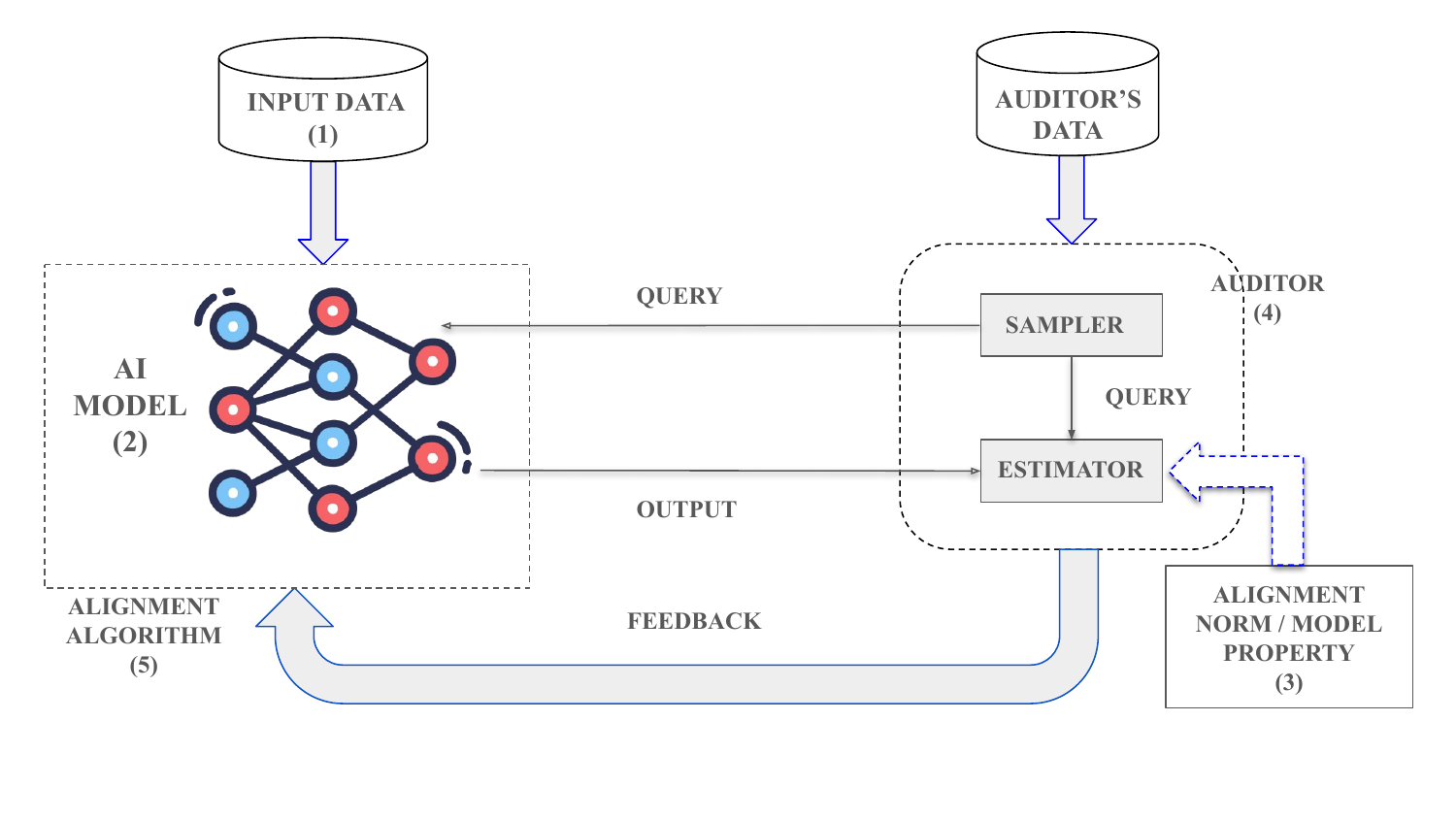}
	\caption{Components of an auditing to alignment mechanism for an AI algorithms.}\label{fig:auditing_game}
\end{figure}

\paragraph{Fair Game: An Overview.} In Figure~\ref{fig:auditing_game}, we illustrate the pipeline for a auditing to alignment feedback mechanism for any AI algorithm. 
First, an input dataset (Component (1)) is used to train an AI Algorithm (Component (2)). This AI algorithm exhibits different alignment norms, also called model properties (Component (3)), such as privacy leakage, bias, and instability (lack of robustness). An auditor (Component (4)) aims to accurately estimate the desired property (or properties) with minimal samples from a data pool, which might or might not match the input dataset depending on the degree of access available to the auditor. 
Then, the deployed auditor sends this feedback to the AI algorithm under audit, which is hardly used to incrementally debias the algorithm at present. Finally, we deploy another alignment algorithm (Component (5)) that leverages the feedback and other side information (e.g. preferences over outcomes), if available, to efficiently socially align the properties of the AI algorithm under audit. 

In present literature, all of these components are assumed to be static over time. But as an application acquires new data over time, updates its model, and socially acceptable ethical norms and regulations also evolve, it motivates the conceptualisation of dynamic auditors. 
This also allows to bring in novel alignment properties from ethics and social sciences if they are computable or estimatable from observable data or their causal relations. 
This flexibility is essential as we still see plethora of robustness and bias metrics to emerge after a decade of studying them, and this is a natural phenomenon as ethics evolve over time and we cannot know beforehand all the impacts of a young and blossoming technology like AI.

\paragraph{Fair Game: Mathematical Formulation.} Let us consider the setting similar to Section~\ref{sec:debias}, i.e. we have a binary classification model trained on a dataset $\traindata \triangleq \{(\mathbf{x}_i, \mathbf{a}_i, y_i)\}_{i=1}^m$, i.e. triplets of non-sensitive features, sensitive features, and outputs, generated from an underlying distribution $\mathcal{D}$. A model trained by minimising the average loss is denoted by $f_{\theta^*}$. Given a measure of bias $\mu$, $f_{\theta^*}$ exhibits a bias $\mu(f_{\theta^*}, \mathcal{D}) \geq 0$.

Now, let us consider that the underlying distribution changes with time $t \in \{1,2,\ldots\}$. Thus, we denote the data distribution, the training dataset, the model, and the property at time $t$ as $\mathcal{D}_t$, $\traindata_t$, $f_{\theta^*,t}$ and $\mu_t$ respectively. This structure defines the first three components in the Fair Game framework (Figure~\ref{fig:auditing_game}).

Under this dynamic setting, we first define \textbf{an anytime-accurate PAC auditor of bias} (Component (4)). The intuition is that an anytime-accurate PAC auditor of bias can achieve below $\epsilon$ error to estimate the desired bias measure as it evolves over time. 
\begin{definition}[Anytime-accurate PAC Auditor]
    An auditor $\mathcal{A}$ is an \textit{anytime-accurate PAC auditor} if for any $\epsilon, \delta \in (0,1)$, there exists a function $m(\epsilon, \delta)$ such that $\forall m \geq m(\epsilon, \delta)$ samples drawn from $\mathcal{D}$, it outputs a mean estimate $\hat{\mu}_{m,t}$ at anytime $t$ satisfying 
    \begin{align}
        \prob( \forall t\in \{1,2,\ldots\}, |\hat{\mu}_{m,t} - \mu_t| \leq \epsilon) \geq 1- \delta\,.
    \end{align}
    Here, the probability is taken over all the stochastic dynamics of the data, the model, and the auditing algorithm, if it uses randomised components. 
\end{definition}
Definition 6 means that with probability $1-\delta$, an anytime-accurate PAC Auditor yields an $\epsilon$-accurate estimate of the property $\mu_t$ exhibited by the model at time $t$.

Now, we define the \textbf{dynamic debiasing algorithm}, which is the final component (Component (5)) required in Fair Game. 
\begin{definition}[Dynamic Debiasing Algorithm]
    Given access to the data distribution, the training dataset, the model, and an estimate of the property at time $t$, i.e. $\mathcal{D}_t$, $\traindata_t$, $f_{\theta^*,t}$ and $\hat{\mu}_t$, 
    a Dynamic Debiasing algorithm $\mech$ minimises the average bias over a given horizon $T \geq 1$, i.e.
    \begin{align}
       V_T(\mech) \triangleq \frac{1}{T} \expect\left[\sum_{t=1}^T \hat{\mu}_t(\mech(f_{\theta^*,t}), \mech(\mathcal{D}_t)) \right]\,.
    \end{align}
    Here, the expectation is taken over all the stochastic dynamics of the data, the model, the bias estimate, and the debiasing algorithm, if it uses randomised components. 
\end{definition}
As the average bias $V_T(\mech)$ of the dynamic debiasing algorithm tends to zero with increase in $T$, it implies that it is able to remove over time the bias in model predictions under the dynamic setup. In general, lower is the $V_T(\mech)$ better is the dynamic debiasing algorithm. 
From reinforcement learning perspective, $V_T(\mech)$ is the value function measuring badness of the debiasing algorithm over time $T$, and the bias exhibited by it at time $t$, i.e., $\hat{\mu}_t(\mech(f_{\theta^*,t}), \mech(\mathcal{D}_t))$, is its cost function per-step.

Finally, with all these components, now we can formally define the Fair Game and its quantitative goals. 
\begin{definition}[Fair Game]\label{def:fair_game}
    Given access to the data distribution, the training dataset, the model, and the property at any time $t$, i.e. $\mathcal{D}_t$, $\traindata_t$, $f_{\theta^*,t}$ and ${\mu}_t$, the auditor-debiasing pair $(\alg,\mech)$ plays a Fair Game by yielding anytime-accurate PAC estimates of the bias, i.e. $\{\hat{\mu,t}\}_{t=1}^T$, and minimising the average bias over time, i.e. $V_T(\mech)$, respectively. 

    We further define the \textbf{regret of the Fair Game} with an auditor-debiasing pair $(\alg,\mech)$ as
    \begin{align}
        \frac{1}{T} \expect\left[\sum_{t=1}^T \hat{\mu}_t(\mech(f_{\theta^*,t}), \mech(\mathcal{D}_t)) \right]-\min_{\alg,\mech} \frac{1}{T} \expect\left[\sum_{t=1}^T {\mu}_t(\mech(f_{\theta^*,t}), \mech(\mathcal{D}_t)) \right] \,.
    \end{align}
\end{definition}
Regret of the Fair Game is the difference between the minimum bias achievable over time by any auditing and debiasing algorithm-pair and that achieved by a deployed system in practice for a given stream of datasets.
We observe that lower regret of the Fair Game indicates higher efficiency of the auditor-debiasing pair $(\alg,\mech)$. \textit{Thus, given a dynamic dataset, an adaptive training algorithm, and the evolving measures of bias, the goal of a Fair Game is to minimise its regret, while deploying anytime-accurate PAC auditors of bias and dynamic debiasing algorithms as two players with interactions.} 
\section{Challenges and Opportunities to Address the Fair Game}\label{sec:properties}
Now, we summarise the four desired properties of the Fair Game framework to audit and debias ML algorithms over time. 

\subsection{\textit{Data Frugality and Accuracy}}
The first pillar of the Fair Game is an anytime-accurate PAC auditor. But the auditor needs to query each of the updated models to conduct the estimation procedure. For external auditors and researchers designing algorithms, the data and access to proprietary ML models become the main bottleneck. \cite{caton2024fairness} mentions this dilemma as

\begin{center}
    ``\textit{This is a hard problem to solve: Companies cannot simply hand out data to researchers, and researchers cannot fix this problem on their own. There is a tension here between advancing the fairness state-of-the-art, privacy, and policy.}"
\end{center}
Thus, auditing over time brings us to the other pole of the AI world, where millions of datapoints are not available at all and we have to sharpen our statistical techniques to collect only informative data leading to accurate estimates. This leads to the first challenge in the Fair game. 

\noindent\textbf{Challenge 1.} \textit{Designing auditors that can use as minimum number of samples to yield as accurate estimate of bias as possible over dynamic data distributions and models. }

Most sequential estimation and  large-scale RL algorithms are known to be ``data-greedy", which is the natural framework for auditing over time. This poses an opportunity to revisit the limits of statistical RL theory in the context of auditing over time as sample frugality becomes imperative. 


\subsection{\textit{Manipulation Proof}}
Manipulation proof is an interesting and unique requirement of an auditor. Specially, an auditing mechanism is a top-down phenomenon in present AI technology scenario where ML models are changing in every economic quarter yielding more profit while we know little about their impacts in socioeconomic, cultural and personal lives. Manipulation proof auditing is specifically important due to two reasons.

\begin{enumerate}
    \item \textit{Robustness to Adversarial Feedback.} To avoid being exposed or fined under auditing and the regulation hammer, a company can provide selected samples to the auditor which make them look fairer. This provides a partial view of the prediction distribution while not being too far from the true one~\citep{yan2022active,godinot2024under}.  
    \item \textit{Opportunity to Evolve.} On the other hand, in the reckoning market of AI, a company might argue that they have to update their models `fast' to stay competitive. Thus, it is fair to give them an opportunity to change their models between two audits~\citep{ajarra2024active}. This provides another motivation to design manipulation-proof auditors that can lead to easy acceptance of auditors in practice.
\end{enumerate}
At this vantage point, we define PAC Auditing with Manipulation-proof Certification that encompasses both the motivations. 
\begin{definition}[PAC Auditing with Manipulation-proof Certification]\label{def:auditype2mp}
    For all $\epsilon, \delta \in (0,1)$, there exist a function $m: (0,1) \to \mathbb{N}$, such that for any probability measure $\mathcal{D} \in \mathcal{P}$, if $S$ is a sample of size $m \geq m(\epsilon, \delta)$ sampled from $\mathcal{D}$, a PAC auditor with manipulation-proof certification yields 
    \begin{itemize}
        \item \textbf{A correct estimate $\hat{\mu}_{m}$:}
    \begin{align*}
        {\prob}  \left[ \sup_{h \in  \mathcal{H}_{\epsilon}(f_\theta^*)} |\hat{\mu}_{m}(h) - {\mu}(h,\mathcal{D})|  \geq \epsilon \right] \leq \delta
    \end{align*}
    \item \textbf{A manipulation-proof region $\mathcal{H}_{\epsilon}(f_\theta^*)$:} 
    \begin{align*}
        {\prob}  \left[ \inf_{h \in {\mathcal{H}_{\epsilon}^C (f_\theta^*)}}  |\hat{\mu}_{m}(h) - {\mu}(h,\mathcal{D})|  \leq \epsilon \right] \leq \delta
    \end{align*}
    \end{itemize}   
    $\mathcal{H}_{\epsilon} (f_\theta^*)$ is a set of models with predictive distributions close to that of $f_\theta^*$, and $\mathcal{H}_{\epsilon}^C (f_\theta^*)$ is its complement. 
\end{definition}
The manipulation-proof region $\mathcal{H}_{\epsilon} (f_\theta^*)$ allows the company under audit to change their models in regulated region around the present model. Manipulation-proofness aims to ensure whether the AI model owner behaves adversarially and provides biased or obsfucated samples, or asks for flexibility to update the model in-between two audits, the auditor should be able to estimate the bias robustly. An obfuscation of the samples or biasing them can be seen as a shift in the prediction distribution of the AI model under audit. Definition 9 claims that if we are sampling from any prediction distribution inside the manipulation-proof region $\mathcal{H}_{\epsilon}^C (f_\theta^*)$ around the true distribution under audit, the auditor can still yield good estimates of bias with high probability. In simple terms, the auditor is manipulation-proof in a regulated region around the true prediction distribution.

For the auditor, it provides an additional constraint, i.e. a region in which its bias estimate would vary minimally due to changes in predictions and input data. 
This goal is often in tension with accurate PAC estimation leading to a tension that we classically observe while designing any accurate by robust estimators~\citep{huber1981robust}.

This also makes PAC auditing with manipulation-proof certification a harder problem than PAC auditing. \cite{yan2022active} show an active learning based procedure to achieve manipulation-proof auditing, while \cite{godinot2024under} show that manipulation-proof auditing can be harder as ML models gets larger and non-linear for a complex dataset. \cite{ajarra2024active} further show that if we use Fourier expansions of prediction distributions for auditing, we by default achieve manipulation-proofness with respect to changes in the smallest one-fourth coefficients. But obtaining a universal complexity measure to quantify hardness of manipulation-proof auditing and comparing it with hardness of classical auditing still remains an open problem. At this point, auditing dynamic algorithms bring a stronger challenge. 

\noindent\textbf{Challenge 2.} \textit{Designing manipulation-proof PAC auditors that can be accurate while computationally efficiently finds the manipulation-proof regions around evolving models over dynamic data distributions and models.}

\subsection{\textit{Adaptive and Dynamic}}
In Section~\ref{sec:framework}, we propose the formal framework of Fair Game. Specifically, Definition~\ref{def:fair_game} formulates it rigorously as a two-player stochastic game. 
Thus, we propose to use the RL for stochastic game (Section~\ref{sec:rl}) as the learning paradigm to resolve the Fair Game efficiently. 

Specifically, last decade has seen a rise in responsible RL that aim to rigorously define and ensure privacy, unbiasedness, and robustness along with utility over time. 
\textbf{Bias in RL} is studied as socio-political and economic policies (whether affirmative or punitive) interact with our society like the policies in RL do (e.g. \texttt{AIRecruiter}, college admissions over years etc.).  
Thus, ensuring fairness in RL posits additional interesting and real-life problems~\citep{buening2021meritocracy}. Researchers have studied effects of different fairness metrics on RL's performance and designed efficient algorithms to tackle them~\citep{gajane2022survey}. But the RL for two-player stochastic games still remains an open problem outside the worst-case and structure-oblivious scenarios. This brings us to the third challenge. 

\noindent\textbf{Challenge 3.} \textit{Designing RL algorithms for two-player games with auditor-debiasing algorithm pairs deployed around evolving models over dynamic data distributions and models.} 

An opportunity arises from the growing study of RL and sequential estimation under constraints. Specifically, we note that \textit{minimising bias over time with auditor feedback of bias is a special case of RL under constraints}. \citep{carlsson:hal-04203235,das2024learning} have derived lower bounds on performance that show how the optimal performance of an RL algorithm depends on the geometry of these constraints. \citep{das2024learning} have reinforced the estimators constraint violations to achieve optimal performance. But these algorithms are still limited to the case of structure-oblivious and linear bandits. It is a scientific opportunity and challenge to extend them further to the Fair Game.

\subsection{\textit{Structured and Preferential Feedback}}
With the growing real-life application of AI, it has become imperative to ensure their behaviour aligned with social norms and users' expectations. Especially, incorporating human preferential feedback in learning (or fine-tuning) process plays a vital role in aligning outputs from a Large Language Model (LLM) socially~\citep{conitzerposition,Preference_collapse, cultural_bias, safe_rlhf}. Reinforcement Learning with Human Feedback (RLHF) enhances this alignment by using human judgments to fine-tune models, guiding them toward preferred actions and responses resulting in better model-adaptivity~\citep{ouyang2022training,christiano2017deep,song2024preference,Genderalign}. But RLHF can lead to over-optimisation for specific preferences, causing models to be overly specialised or biased, which repels adaptivity to diverse, unseen preferences in real-world applications~\citep{christiano2017deep,ziegler2019fine}. In this context, ~\cite{shukla:hal-04733134} propose the first preference-dependent lower bounds for bandits with multiple objectives and incomplete preferences. Even in this simpler setting, we observe that the preferences distort decision space.  {\textit{But we understand very little how the incomplete preferences and high-dimensional features on continuous state-action spaces distort the decision space, which are closer to the LLMs.}
Thus, the fourth challenge comes as follows.

\noindent\textbf{Challenge 4.} \textit{Designing debiasing algorithms that can optimally incorporate preferential and qualitative feedback of auditors to better debias the ML models over time.}

\subsection{\textit{Final Destination: Existence of Equilibrium?}}
The final question in any game is the existence of an \textbf{equilibrium}. 
We observe that bias measures can be PAC audited sample-efficiently with a universal auditor~\citep{ajarra2024active}, while one can consider debiasing as a constraint optimisation problem of minimising loss while keeping the bias upper bounded. Thus, in presence of an auditor's feedback that accurately quantifies the bias and instability of a dynamic algorithm at any point of time, we can treat minimisation of bias and instability in a dynamic AI algorithm under constraints on the prediction distribution. 
This strategy has been studied in offline setting as Empirical Risk Minimisation (ERM) with distributional constraints. In addition, we know that the constraint violations can be used in feedback with a dynamic learning algorithm to achieve the desired safety and unbiasedness over time~\citep{fletberliac:hal-03771734}. 
The generic framework to address them is to simulate a constraint-breaking adversary and a learner trying to avoid the adversary by only looking into its feedback. They use the same data stream to conduct their learning procedures. 
This poses the final two challenges.

\noindent\textbf{Challenge 5.} \textit{Can we achieve a stable equilibrium for the Fair Game when the measure of bias is fixed over time?}

\noindent\textbf{Challenge 6.} \textit{Can we achieve a stable equilibrium for the Fair Game when the measure of bias is evolving over time?}

We conjecture that the answer to the first challenge is affirmative while that of the second one depends on the changes in bias measures. 
Our intuition is based on the fact that bias measures are often conflicting and thus, cannot be achieved simultaneously in a single game. 
One avenue to address these problems will be to extend the growing literature on stochastic non-zero-sum games~\citep{sorin1986asymptotic,zhang2020model,fiegel2023adapting,bai2022near} with learners to the Fair Game setting, where utility of the auditor is to measure the constraint violation and that of the learner is to minimise loss in training while incorporating the auditor's feedback. 
\section{Bridging Fair Game and Legal Perspectives of Audits}
As Fair Game aims to bind the auditor and model owner into a single framework, it is a natural requirement to develop a legal framework for this, and wonder how the present legal frameworks for auditing do or do not satisfy the requirements. 

\paragraph{The Landscape of Law and Algorithmic Audits.} \cite{le2023algorithmic} provide an overview of the existing legal intricacies around algorithmic audits. We extend their insights to formalise the desired legal bindings for Fair Game. 

The first legal concern for algorithmic auditing is data protection laws, such as the General Data Protection Regulation (GDPR)~\citep{gdpr} in Europe. These laws safeguard user privacy and restrict data access to the algorithmic auditors (\textit{Directive
96/9/EC of the European Parliament and of the Council of 11 March 1996
on the Legal Protection of Databases 2019}). In this context, \cite{le2023algorithmic} distinguishes between two types of audits: Bobby Audits, which use real user data and thus face stronger legal scrutiny, and Sherlock Audits, which rely on synthetic data and are less legally restrictive but may have weaker evidential value. Fair Game naturally inherits these challenges of static auditing evoked by data protection laws. In this context, the rich literature of estimation and learning under privacy would play an instrumental role to develop efficient technical solutions~\citep{dpbook,dperm,dpregression,wang2020dp}. 

Another challenge in algorithmic auditing is due to intellectual property rights and trade secrets. Many companies argue that their algorithms are proprietary, and in turn, constraint the auditors access of information to examine them. The present laws currently does not provide a unequivocal ``right to audit". Thus, auditors often have to rely on indirect methods, such as scrapping the data from web services and digital platforms, or risk potential legal consequences, if they test proprietary algorithms without explicit permission. This propels the development of black-box auditors for estimating bias of AI models and is an active area of research~\citep{ghosh2023biased,ajarra2024active}. This nuance is naturally covered by the Fair Game framework as it is applicable to both internal and external auditors with white-box and black-box access, respectively.  

Liability is another challenge\footnote{For example, Article 9 of French Civil Code states that ``Each party has the burden of proving in accordance with the law the facts necessary for the success of its claim."}. If auditors publicly disclose biases or discriminatory practices within an algorithm, they may face legal threats from companies seeking to protect their reputation and avoid regulatory penalties~\citep{le2023algorithmic}. Additionally, how can we ensure that the companies would provide the auditors unbiased access to samples, or update their models inside the prescribed manipulation-proof regions. Right now there is no reward or restriction on companies to not play adversarial during an audit. Thus, a legal binding of trust and liabilities between companies and auditors to create one single ecosystem is presently missing. Such developments has been seen for some other technologies, such as ISO\footnote{\url{https://www.iso.org/standards.html}} and IEEE\footnote{\url{https://standards.ieee.org/ieee/24774/10126/}} standards and certifications, and we need to develop one for emerging AI technologies. 
A unifying legal framework would be imperative to turn Fair Game into an effective paradigm for trustworthy deployment of AI.

The legal environment for algorithmic auditing is still evolving. While current laws protect companies and user privacy, they must also facilitate responsible and ethical auditing. We now discuss about NYC Bias Audit Law, which is one of the real-life instance in this direction. 

\paragraph{A Bias Auditing Law in Real-life: NYC Bias Audit Law.} As a real-life example of law enforcing algorithmic audits, we discuss the NYC Bias Audit Law or the Local Law 144 that has been mandated for algorithmic auditing of AI driven employment tools called ``AEDT"s (Automated Employment Decision Tool)\footnote{AEDTs are real-life instances of our conceptual algorithm \texttt{AIRecruiter}.}.  This is one of the first regulation enforced in the United States (specifically, New York City) that proposed auditing of digital platforms. Enacted in 2021 and enforced as of July 5, 2023, Local Law 144 is a significant step towards fairness and transparency in automated decision making, particularly targeting employment and promotion processes. 

\noindent\textit{Key Components.} 1. \textit{Annual Independent Audits of Bias.} Potential employers or organisations must conduct an independent bias audit of their AEDTs at least once per year. The audit evaluates bias in predictions across the protected demographics, specifically race, ethnicity, and gender. The fairness metric used in auditing AEDTs is the ratio 
\begin{align*}
    \frac{\text{Selection rate for a category}}{\text{Selection rate of the most selected category}}\,.
\end{align*}
This is a case-specific estimate of demographic parity (Definition 2) studied in fair ML literature. As the ratio goes to one, the AEDT is considered to be more fair.

2. \textit{Candidate and Employee Notification.}  Any employer must notify job applicants and employees at least 10 business days before using an AEDT in decision-making. It must include information about the tool's functionality and the applicant's right to request alternative evaluation methods. This aims to ensure transparency and privacy-compliant data processing in the AEDT tool.

3. \textit{Public Disclosure of Audit Results.} Any user of AEDTs must publicly post the results of their annual bias audits in their website demonstrating the system's fairness and impact on different protected demographics. This ensures the liability of AEDTs as a socially impactful technology as discussed before.

4. \textit{Penalties for Non-Compliance.} Organisations failing to comply with these requirements face financial penalties, starting at $500$ dollars for the first violation and increasing for subsequent infractions. This component aims to reinforce compliance of the user to the law and ensure unbiasedness in employment. 

\noindent\textit{A Failed Attempt or a Case for Unified Dynamic Auditing?} 
Despite of being a pioneering attempt towards bridging theoretical auditing and practical enforcement as a legal obligation for employers, the Local Law 144 have had many shortcomings.

\begin{enumerate}
    \item \textit{Simplistic Auditing Process.} It is \textit{too optimistic} that the auditor can use the final output and compute one of the many group fairness metrics to measure bias in order to create a comprehensive bias report. It does not promote for any causation of the bias and also no preference-based or case-to-case evaluation. It is very hard to develop any legal case with such simplistic audit. Fair Game brings in a richer framework to mitigate these issues. 
    \item \textit{Static Auditing.} The auditing approach, specially the fairness metric suggested by the law, is not adaptive with time. It could neither leverage historical information in the auditing process nor can keep up with the changing notions of fairness. This is problematic as we cannot understand whether an AEDT tool is systematically biased or it is just one exceptional event. At this point, Fair Game brings in the dynamic perspective and proposes to treat the auditing and debiasing as a two player game, where both players have different incentives to play the game. 
    \item \textit{Lack of Enforcing Compliance.} Only one case of non-compliance was officially reported in six months after commencement though the authority was aware of hundreds of such cases. Also, there was no demand for third party auditor services, like \href{https://proceptual.com/}{proceptual.com}~\citep{janaro2023nyc,wright2024null}. The reasons are two-fold. \textit{First}, bureaucracy and lobbying delayed the enforcement of the law. Even after imposing it, NY city council loosened the laws and requirements and left several loops in the compliance rules within 24 hours of commencement. There is no purely technical solution to this. \textit{Second}, the present law neither provides any reward or pressing reason for the AI model owner to comply with the auditing law, nor it gives any flexibility and guideline to update the AI models between two audits. Fair Game proposes the dynamic and adaptive framework to overcome these issues.
\end{enumerate}

\section{Discussions and Perspectives}\label{sec:conclusion}
With the growing use of AI algorithms for bouquets of real-life tasks such as autonomous recruitment, crime recidivism, autonomous college admission etc., the need for debiasing these algorithms from any bias and aligning them with social norms has become crucial. Throughout this article, we have explained different sources of biases with real-life examples like AEDTs as well as the different approaches taken to mitigate them. We propose a dynamic mechanism ``Fair Game" that assures to preserve fairness in prediction of ML algorithms, that also adapt to changes as the society interacts with the algorithm over time.  The ``Fair Game" conjoins an \textit{Auditor} and a \textit{Debiasing Algorithm} in a loop leveraging the concepts of Reinforcement Learning. Our approach addresses the lacunae of the existing approaches to debias an ML algorithm as they usually work well for static environments. ``Fair Game" takes the process of debiasing to a dynamic environment that makes the process adaptive over time. We have also emphasised on some desirable properties of an auditor such as accuracy, data frugality, constant manipulation proofness etc. Thus, ``Fair Game" envisions an interactive approach to design statistically and computationally efficient continual auditors of AI algorithms, and incorporate auditor's feedback in-a-loop to dynamically align these algorithms. This approach is novel, descriptive, more regulation-friendly, and allows continual evaluation and alignment of AI algorithms.    

\paragraph{Collaborative Human and Algorithmic Auditing.} Finally, we would like to emphasise that the Fair Game does not aim to replace but aid the human AI auditors. This is necessary in order to connect the algorithmic audits with the available legal frameworks. Specifically, Fair Game enables the human auditors in two ways.

(i) For companies dealing with large number of user data, for example AIRecruiter, an auditor will have to manage and analyse a big and dynamic database. In these scenarios, a human-driven model becomes challenging in terms of resources and capacity. The AI Auditor rather can provide statistically sound summaries of bias and other risks due to deployment of the AI algorithm that the human auditors and lawyers can use for further inspection. 

(ii) In Fair Game, if an auditor is given a task to align an AI algorithm as per the requirements in DMA or DSA (for example, services provided by \href{babl.ai}{babl.ai}), it demands intrinsic knowledge of the specific domain of application. This is where algorithmic auditors can enable us significantly by saving resources and creating a knowledge map that can be adapted over time. 

But algorithms do not understand ethics till we design and tune them to it. In recent days, we have witnessed numerous proposals and research articles that advocates human intervention in aligning LLMs with social values in their decision making whether by prompt engineering or leveraging RL with human feedback (RLHF)~\citep{wang2023aligning,shankar2024validates,ji2024beavertails,song2024preference}.  Thus, we envision the use of structural and preferential feedback (Section 4.4) in Fair Game, which allows human auditors to intervene with case studies and preference over certain outcomes, and then the algorithmic auditor and the alignment algorithm further adapts to it.
\section*{Acknowledgments}
We acknowledge the ANR JCJC project REPUBLIC (ANR-22-CE23-0003-01), the PEPR project FOUNDRY (ANR23-PEIA-0003), and the Regalia Project partnered by Inria and French Ministry of Finance.
\bibliographystyle{apalike}
\bibliography{ref}

\begin{thebibliography}{}

\bibitem[Agarwal et~al., 2018]{agarwal2018reductions}
Agarwal, A., Beygelzimer, A., Dud{\'\i}k, M., Langford, J., and Wallach, H.
  (2018).
\newblock A reductions approach to fair classification.
\newblock In {\em International conference on machine learning}, pages 60--69.
  PMLR.

\bibitem[Ajarra et~al., 2024]{ajarra2024active}
Ajarra, A., Ghosh, B., and Basu, D. (2024).
\newblock Active fourier auditor for estimating distributional properties of ml
  models.
\newblock {\em arXiv preprint arXiv:2410.08111}.

\bibitem[Albarghouthi et~al., 2017]{albarghouthi2017fairsquare}
Albarghouthi, A., D'Antoni, L., Drews, S., and Nori, A.~V. (2017).
\newblock Fairsquare: probabilistic verification of program fairness.
\newblock {\em Proc. ACM Program. Lang.}, 1(OOPSLA).

\bibitem[Altman, 1999]{altman1999constrained}
Altman, E. (1999).
\newblock {\em
  \href{https://hal.inria.fr/docs/00/07/41/09/PS/RR-2574.ps}{Constrained Markov
  decision processes}}, volume~7.
\newblock CRC Press.

\bibitem[Angwin et~al., 2016]{compas}
Angwin, J., Larson, J., Mattu, S., and Kirchner, L. (23 May 2016).
\newblock
  \href{http://www.propublica.org/article/machine-bias-risk-assessments-in-criminal-sentencing}{Machine
  bias: There’s software used across the country to predict future criminals.
  And it’s biased against blacks,} \textit{ProPublica}.

\bibitem[Annas, 2003]{hipaa}
Annas, G.~J. (2003).
\newblock Hipaa regulations: a new era of medical-record privacy?
\newblock {\em New England Journal of Medicine}, 348:1486.

\bibitem[Arrow, 1971]{arrow}
Arrow, K. (1971).
\newblock The theory of discrimination.

\bibitem[Assaf et~al., 2020]{assaf2020utilization}
Assaf, D., Gutman, Y., Neuman, Y., Segal, G., Amit, S., Gefen-Halevi, S.,
  Shilo, N., Epstein, A., Mor-Cohen, R., Biber, A., et~al. (2020).
\newblock Utilization of machine-learning models to accurately predict the risk
  for critical covid-19.
\newblock {\em Internal and emergency medicine}, 15:1435--1443.

\bibitem[Azize and Basu, 2022]{azize:hal-03781600}
Azize, A. and Basu, D. (2022).
\newblock {When Privacy Meets Partial Information: A Refined Analysis of
  Differentially Private Bandits}.
\newblock In {\em {Advances in Neural Information Processing Systems}}, New
  Orleans, United States.

\bibitem[Azize and Basu, 2024]{azize:hal-04611650}
Azize, A. and Basu, D. (2024).
\newblock {Concentrated Differential Privacy for Bandits}.
\newblock In {\em {2024 IEEE Conference on Secure and Trustworthy Machine
  Learning (SaTML)}}, pages 78--109, Toronto, Canada. {IEEE}, {IEEE}.

\bibitem[Azize et~al., 2023]{azize:hal-04215474}
Azize, A., Jourdan, M., Marjani, A.~A., and Basu, D. (2023).
\newblock {On the Complexity of Differentially Private Best-Arm Identification
  with Fixed Confidence}.
\newblock In {\em {NeurIPS 2023 -- Conference on Neural Information Processing
  Systems}}, volume~36, pages 71150--71194, New Orleans (US), United States.

\bibitem[Bagaric et~al., 2019]{bagaric2019erasing}
Bagaric, M., Hunter, D., and Stobbs, N. (2019).
\newblock Erasing the bias against using artificial intelligence to predict
  future criminality: algorithms are color blind and never tire.
\newblock {\em U. Cin. L. Rev.}, 88:1037.

\bibitem[Bai et~al., 2022]{bai2022near}
Bai, Y., Jin, C., Mei, S., and Yu, T. (2022).
\newblock Near-optimal learning of extensive-form games with imperfect
  information.
\newblock In {\em International Conference on Machine Learning}, pages
  1337--1382. PMLR.

\bibitem[Barocas et~al., 2023]{barocas2023fairness}
Barocas, S., Hardt, M., and Narayanan, A. (2023).
\newblock {\em Fairness and machine learning: Limitations and opportunities}.
\newblock MIT Press.

\bibitem[Bastani et~al., 2019]{bastani2019probabilistic}
Bastani, O., Zhang, X., and Solar-Lezama, A. (2019).
\newblock Probabilistic verification of fairness properties via concentration.
\newblock {\em Proceedings of the ACM on Programming Languages},
  3(OOPSLA):1--27.

\bibitem[Basu et~al., 2020]{basu2019differential}
Basu, D., Dimitrakakis, C., and Tossou, A. (2020).
\newblock \href{https://arxiv.org/pdf/1905.12298.pdf}{Privacy in Multi-armed
  Bandits: Fundamental Definitions and Lower Bounds}.
\newblock {\em NeurIPS Workshop on Privacy Preserving Machine Learning}.

\bibitem[Basu et~al., 2022]{basu2022bandits}
Basu, D., Maillard, O.-A., and Mathieu, T. (2022).
\newblock \href{https://arxiv.org/pdf/2203.03186.pdf}{Bandits corrupted by
  Nature: Lower bounds on regret and robust optimistic algorithm}.
\newblock {\em arXiv preprint arXiv:2203.03186}.
\newblock (Submitted to COLT'22).

\bibitem[Bellamy et~al., 2019]{bellamy2019ai}
Bellamy, R.~K., Dey, K., Hind, M., Hoffman, S.~C., Houde, S., Kannan, K.,
  Lohia, P., Martino, J., Mehta, S., Mojsilovi{\'c}, A., et~al. (2019).
\newblock Ai fairness 360: An extensible toolkit for detecting and mitigating
  algorithmic bias.
\newblock {\em IBM Journal of Research and Development}, 63(4/5):4--1.

\bibitem[Bellamy et~al., 2018]{aif360-oct-2018}
Bellamy, R. K.~E., Dey, K., Hind, M., Hoffman, S.~C., Houde, S., Kannan, K.,
  Lohia, P., Martino, J., Mehta, S., Mojsilovic, A., Nagar, S., Ramamurthy,
  K.~N., Richards, J., Saha, D., Sattigeri, P., Singh, M., Varshney, K.~R., and
  Zhang, Y. (2018).
\newblock Ai fairness 360: An extensible toolkit for detecting, understanding,
  and mitigating unwanted algorithmic bias.

\bibitem[Blumrosen, 1967]{blumrosen1967duty}
Blumrosen, A.~W. (1967).
\newblock The duty of fair recruitment under the civil rights act of 1964.
\newblock {\em Rutgers L. Rev.}, 22:465.

\bibitem[Borca-Tasciuc et~al., 2022]{borca2022provable}
Borca-Tasciuc, G., Guo, X., Bak, S., and Skiena, S. (2022).
\newblock Provable fairness for neural network models using formal
  verification.

\bibitem[Brown, 1951]{brown1951iterative}
Brown, G.~W. (1951).
\newblock Iterative solution of games by fictitious play.
\newblock {\em Act. Anal. Prod Allocation}, 13(1):374.

\bibitem[Buening et~al., 2022]{buening:hal-03445971}
Buening, T.~K., Segal, M., Basu, D., George, A.-M., and Dimitrakakis, C.
  (2022).
\newblock {On Meritocracy in Optimal Set Selection}.
\newblock In {\em {EAAMO 2022- Equity and Access in Algorithms, Mechanisms, and
  Optimization}}, Arlington, United States. {ACM}.

\bibitem[Bénesse et~al., 2021]{benesse2021}
Bénesse, C., Gamboa, F., Loubes, J.-M., and Boissin, T. (2021).
\newblock Fairness seen as global sensitivity analysis.

\bibitem[Calders and {\v{Z}}liobait{\.e}, 2013]{calders2013unbiased}
Calders, T. and {\v{Z}}liobait{\.e}, I. (2013).
\newblock Why unbiased computational processes can lead to discriminative
  decision procedures.
\newblock In {\em Discrimination and Privacy in the Information Society: Data
  mining and profiling in large databases}, pages 43--57. Springer.

\bibitem[Carlsson et~al., 2024]{carlsson:hal-04203235}
Carlsson, E., Basu, D., Johansson, F.~D., and Dubhashi, D. (2024).
\newblock {Pure Exploration in Bandits with Linear Constraints}.
\newblock In {\em {International Conference on Artificial Intelligence and
  Statistics}}, volume 238 of {\em Proceedings of Machine Learning Research
  (PMLR)}, pages 334--342, Valencia (Espagne), Spain.

\bibitem[cas et~al., 2023]{fairMLbook}
cas, S., Hardt, M., and Narayanan, A. (2023).
\newblock {\em Fairness and Machine Learning: Limitations and Opportunities}.
\newblock MIT Press.

\bibitem[Caton and Haas, 2024]{caton2024fairness}
Caton, S. and Haas, C. (2024).
\newblock Fairness in machine learning: A survey.
\newblock {\em ACM Computing Surveys}, 56(7):1--38.

\bibitem[Celis et~al., 2019]{celis2019classification}
Celis, L.~E., Huang, L., Keswani, V., and Vishnoi, N.~K. (2019).
\newblock Classification with fairness constraints: A meta-algorithm with
  provable guarantees.
\newblock In {\em Proceedings of the conference on fairness, accountability,
  and transparency}, pages 319--328.

\bibitem[Chaudhuri et~al., 2011]{dperm}
Chaudhuri, K., Monteleoni, C., and Sarwate, A.~D. (2011).
\newblock Differentially private empirical risk minimization.
\newblock {\em Journal of Machine Learning Research}, 12(3).

\bibitem[Cheng et~al., 2021]{cheng2021socially}
Cheng, L., Varshney, K.~R., and Liu, H. (2021).
\newblock Socially responsible {AI} algorithms: Issues, purposes, and
  challenges.
\newblock {\em Journal of Artificial Intelligence Research}, 71:1137--1181.

\bibitem[Cherian and Cand{{\`e}}s, 2024]{cherian2023statistical}
Cherian, J.~J. and Cand{{\`e}}s, E.~J. (2024).
\newblock Statistical inference for fairness auditing.
\newblock {\em Journal of Machine Learning Research}, 25(149):1--49.

\bibitem[Chierichetti et~al., 2019]{chierichetti2019matroids}
Chierichetti, F., Kumar, R., Lattanzi, S., and Vassilvtiskii, S. (2019).
\newblock Matroids, matchings, and fairness.
\newblock In {\em The 22nd international conference on artificial intelligence
  and statistics}, pages 2212--2220. PMLR.

\bibitem[Chouldechova, 2017]{chouldechova2017fair}
Chouldechova, A. (2017).
\newblock Fair prediction with disparate impact: A study of bias in recidivism
  prediction instruments.
\newblock {\em Big data}, 5(2):153--163.

\bibitem[Chouldechova and Roth, 2018]{chouldechova2018frontiers}
Chouldechova, A. and Roth, A. (2018).
\newblock The frontiers of fairness in machine learning.
\newblock {\em arXiv preprint arXiv:1810.08810}.

\bibitem[Chouldechova and Roth, 2020]{chouldechova2020snapshot}
Chouldechova, A. and Roth, A. (2020).
\newblock \href{https://dl.acm.org/doi/pdf/10.1145/3376898}{A snapshot of the
  frontiers of fairness in machine learning}.
\newblock {\em Communications of the ACM}, 63(5):82--89.

\bibitem[Christiano et~al., 2017]{christiano2017deep}
Christiano, P.~F., Leike, J., Brown, T., Martic, M., Legg, S., and Amodei, D.
  (2017).
\newblock Deep reinforcement learning from human preferences.
\newblock {\em Advances in neural information processing systems}, 30.

\bibitem[Chzhen et~al., 2020]{chzhen2020fair}
Chzhen, E., Denis, C., Hebiri, M., Oneto, L., and Pontil, M. (2020).
\newblock \href{https://arxiv.org/pdf/2006.07286.pdf}{Fair regression with
  Wasserstein barycenters}.
\newblock {\em Advances in Neural Information Processing Systems},
  33:7321--7331.

\bibitem[Conitzer et~al., 2024]{conitzerposition}
Conitzer, V., Freedman, R., Heitzig, J., Holliday, W.~H., Jacobs, B.~M.,
  Lambert, N., Moss{\'e}, M., Pacuit, E., Russell, S., Schoelkopf, H., et~al.
  (2024).
\newblock Position: Social choice should guide ai alignment in dealing with
  diverse human feedback.
\newblock In {\em Forty-first International Conference on Machine Learning}.

\bibitem[Corbett-Davies et~al., 2017]{corbett2017algorithmic}
Corbett-Davies, S., Pierson, E., Feller, A., Goel, S., and Huq, A. (2017).
\newblock Algorithmic decision making and the cost of fairness.
\newblock In {\em Proceedings of the 23rd acm sigkdd international conference
  on knowledge discovery and data mining}, pages 797--806.

\bibitem[Cotter et~al., 2019]{cotter2019optimization}
Cotter, A., Jiang, H., Gupta, M., Wang, S., Narayan, T., You, S., and
  Sridharan, K. (2019).
\newblock Optimization with non-differentiable constraints with applications to
  fairness, recall, churn, and other goals.
\newblock {\em Journal of Machine Learning Research}, 20(172):1--59.

\bibitem[D{{a}}browski and Suska, 2022]{dsadma}
D{{a}}browski, {\L}.~D. and Suska, M. (2022).
\newblock {\em The European Union Digital Single Market: Europe's Digital
  Transformation}.
\newblock Routledge.

\bibitem[Dai et~al., 2023]{safe_rlhf}
Dai, J., Pan, X., Sun, R., Ji, J., Xu, X., Liu, M., Wang, Y., and Yang, Y.
  (2023).
\newblock Safe rlhf: Safe reinforcement learning from human feedback.
\newblock {\em arXiv preprint arXiv:2310.12773}.

\bibitem[Dandekar et~al., 2018]{dpregression}
Dandekar, A., Basu, D., and Bressan, S. (2018).
\newblock Differential privacy for regularised linear regression.
\newblock In {\em International Conference on Database and Expert Systems
  Applications}, pages 483--491. Springer.

\bibitem[Das and Basu, 2024]{das2024learning}
Das, U. and Basu, D. (2024).
\newblock Learning to explore with lagrangians for bandits under unknown
  constraints.
\newblock In {\em Seventeenth European Workshop on Reinforcement Learning}.

\bibitem[Daskalakis et~al., 2023]{daskalakis2023complexity}
Daskalakis, C., Golowich, N., and Zhang, K. (2023).
\newblock The complexity of markov equilibrium in stochastic games.
\newblock In {\em The Thirty Sixth Annual Conference on Learning Theory}, pages
  4180--4234. PMLR.

\bibitem[Devroye et~al., 2013]{devroye2013probabilistic}
Devroye, L., Gy{\"o}rfi, L., and Lugosi, G. (2013).
\newblock {\em A probabilistic theory of pattern recognition}, volume~31.
\newblock Springer Science \& Business Media.

\bibitem[Dwork and Roth, 2014]{dpbook}
Dwork, C. and Roth, A. (2014).
\newblock \href{https://www.tau.ac.il/~saharon/BigData2018/privacybook.pdf}{The
  algorithmic foundations of differential privacy}.
\newblock {\em Found. Trends Theor. Comput. Sci.}, 9(3-4):211--407.

\bibitem[Elie et~al., 2020]{elie2020convergence}
Elie, R., Perolat, J., Lauri{\`e}re, M., Geist, M., and Pietquin, O. (2020).
\newblock On the convergence of model free learning in mean field games.
\newblock In {\em Proceedings of the AAAI Conference on Artificial
  Intelligence}, volume~34, pages 7143--7150.

\bibitem[Eriksson et~al., 2022a]{eriksson:hal-03770369}
Eriksson, H., Basu, D., Alibeigi, M., and Dimitrakakis, C. (2022a).
\newblock {Risk-Sensitive Bayesian Games for Multi-Agent Reinforcement Learning
  under Policy Uncertainty}.
\newblock In {\em {OptLearnMAS@AAMAS}}, Workshop on Optimization and Learning
  in Multiagent Systems at International Conference on Autonomous Agents and
  Multiagent Systems, Virtual, New Zealand.

\bibitem[Eriksson et~al., 2022b]{eriksson:hal-03150823}
Eriksson, H., Basu, D., Alibeigi, M., and Dimitrakakis, C. (2022b).
\newblock {SENTINEL: Taming Uncertainty with Ensemble-based Distributional
  Reinforcement Learning}.
\newblock In {\em {UAI 2022- Proceedings of the Thirty-Eighth Conference on
  Uncertainty in Artificial Intelligence}}, volume 180 of {\em Proceedings of
  Machine Learning Research}, pages 631--640, Eindhoven, Netherlands.

\bibitem[Feldman et~al., 2015]{feldman2015certifying}
Feldman, M., Friedler, S.~A., Moeller, J., Scheidegger, C., and
  Venkatasubramanian, S. (2015).
\newblock Certifying and removing disparate impact.
\newblock In {\em proceedings of the 21th ACM SIGKDD international conference
  on knowledge discovery and data mining}, pages 259--268.

\bibitem[Feldman et~al., 2012]{feldman2012agnostic}
Feldman, V., Guruswami, V., Raghavendra, P., and Wu, Y. (2012).
\newblock Agnostic learning of monomials by halfspaces is hard.
\newblock {\em SIAM Journal on Computing}, 41(6):1558--1590.

\bibitem[Fiegel et~al., 2023]{fiegel2023adapting}
Fiegel, C., M{\'e}nard, P., Kozuno, T., Munos, R., Perchet, V., and Valko, M.
  (2023).
\newblock Adapting to game trees in zero-sum imperfect information games.
\newblock In {\em International Conference on Machine Learning}, pages
  10093--10135. PMLR.

\bibitem[Fiss, 1970]{fiss1970theory}
Fiss, O.~M. (1970).
\newblock A theory of fair employment laws.
\newblock {\em U. Chi. L. Rev.}, 38:235.

\bibitem[Flet-Berliac and Basu, 2022]{fletberliac:hal-03771734}
Flet-Berliac, Y. and Basu, D. (2022).
\newblock {SAAC: Safe Reinforcement Learning as an Adversarial Game of
  Actor-Critics}.
\newblock In {\em {RLDM 2022 - The Multi-disciplinary Conference on
  Reinforcement Learning and Decision Making}}, Providence, United States.
\newblock Accepted at the 5th Multi-disciplinary Conference on Reinforcement
  Learning and Decision Making (RLDM 2022).

\bibitem[Gajane et~al., 2022]{gajane2022survey}
Gajane, P., Saxena, A., Tavakol, M., Fletcher, G., and Pechenizkiy, M. (2022).
\newblock Survey on fair reinforcement learning: Theory and practice.
\newblock {\em arXiv preprint arXiv:2205.10032}.

\bibitem[Galhotra et~al., 2017]{galhotra2017fairness}
Galhotra, S., Brun, Y., and Meliou, A. (2017).
\newblock Fairness testing: testing software for discrimination.
\newblock In {\em Proceedings of the 2017 11th Joint meeting on foundations of
  software engineering}, pages 498--510.

\bibitem[Galindo and Tamayo, 2000]{galindo2000credit}
Galindo, J. and Tamayo, P. (2000).
\newblock Credit risk assessment using statistical and machine learning: basic
  methodology and risk modeling applications.
\newblock {\em Computational economics}, 15:107--143.

\bibitem[Ghosh et~al., 2023a]{ghosh:hal-03770346}
Ghosh, B., Basu, D., and Meel, K. (2023a).
\newblock {''How Biased are Your Features?'': Computing Fairness Influence
  Functions with Global Sensitivity Analysis}.
\newblock In {\em {FAccT '23: the 2023 ACM Conference on Fairness,
  Accountability, and Transparency}}, pages 138--148, Chicago IL, United
  States. {ACM}.

\bibitem[Ghosh et~al., 2021]{ghosh2021justicia}
Ghosh, B., Basu, D., and Meel, K.~S. (2021).
\newblock Justicia: A stochastic sat approach to formally verify fairness.
\newblock In {\em Proceedings of the AAAI Conference on Artificial
  Intelligence}, volume~35, pages 7554--7563.

\bibitem[Ghosh et~al., 2022a]{ghosh2022fvgm}
Ghosh, B., Basu, D., and Meel, K.~S. (2022a).
\newblock Algorithmic fairness verification with graphical models.
\newblock In {\em Proceedings of the AAAI Conference on Artificial
  Intelligence}, volume~36, pages 9539--9548.

\bibitem[Ghosh et~al., 2022b]{ghosh2022algorithmic}
Ghosh, B., Basu, D., and Meel, K.~S. (2022b).
\newblock Algorithmic fairness verification with graphical models.
\newblock In {\em Proceedings of the AAAI Conference on Artificial
  Intelligence}, volume~36, pages 9539--9548.

\bibitem[Ghosh et~al., 2022c]{Ghosh_23}
Ghosh, B., Basu, D., and Meel, K.~S. (2022c).
\newblock {Algorithmic fairness verification with graphical models}.
\newblock In {\em {AAAI-2022 - 36th AAAI Conference on Artificial
  Intelligence}}, volume~2, Virtual, United States.

\bibitem[Ghosh et~al., 2023b]{ghosh2023biased}
Ghosh, B., Basu, D., and Meel, K.~S. (2023b).
\newblock ``how biased are your features?”: Computing fairness influence
  functions with global sensitivity analysis.
\newblock In {\em Proceedings of the 2023 ACM Conference on Fairness,
  Accountability, and Transparency}, pages 138--148.

\bibitem[Giannou et~al., 2022]{giannou2022convergence}
Giannou, A., Lotidis, K., Mertikopoulos, P., and Vlatakis-Gkaragkounis, E.-V.
  (2022).
\newblock On the convergence of policy gradient methods to nash equilibria in
  general stochastic games.
\newblock {\em Advances in Neural Information Processing Systems},
  35:7128--7141.

\bibitem[Godinot et~al., 2023]{godinot}
Godinot, A., Le~Merrer, E., Tr{\'e}dan, G., Penzo, C., and Ta{\"i}ani, F.
  (2023).
\newblock {Change-Relaxed Active Fairness Auditing}.
\newblock In {\em {RJCIA 2023 - 21e Rencontres des Jeunes Chercheurs en
  Intelligence Artificiel}}, CNIA, pages 91--96, Strasbourg, France.
  {Association Fran{\c c}aise pour l'Intelligence Artificielle}.

\bibitem[Godinot et~al., 2024]{godinot2024under}
Godinot, A., Le~Merrer, E., Tr{\'e}dan, G., Penzo, C., and Ta{\"\i}ani, F.
  (2024).
\newblock Under manipulations, are some ai models harder to audit?
\newblock In {\em 2024 IEEE Conference on Secure and Trustworthy Machine
  Learning (SaTML)}, pages 644--664. IEEE.

\bibitem[Goldwasser et~al., 2021]{goldwasser2021interactive}
Goldwasser, S., Rothblum, G.~N., Shafer, J., and Yehudayoff, A. (2021).
\newblock Interactive proofs for verifying machine learning.
\newblock In {\em 12th Innovations in Theoretical Computer Science Conference
  (ITCS 2021)}. Schloss-Dagstuhl-Leibniz Zentrum f{\"u}r Informatik.

\bibitem[Gordaliza et~al., 2019]{gordaliza2019obtaining}
Gordaliza, P., Del~Barrio, E., Fabrice, G., and Loubes, J.-M. (2019).
\newblock Obtaining fairness using optimal transport theory.
\newblock In {\em International conference on machine learning}, pages
  2357--2365. PMLR.

\bibitem[Gorti et~al., 2024]{gorti2024unboxing}
Gorti, A., Gaur, M., and Chadha, A. (2024).
\newblock Unboxing occupational bias: Grounded debiasing llms with us labor
  data.
\newblock {\em arXiv preprint arXiv:2408.11247}.

\bibitem[Gu et~al., 2017]{gu2017deep}
Gu, S., Holly, E., Lillicrap, T., and Levine, S. (2017).
\newblock Deep reinforcement learning for robotic manipulation with
  asynchronous off-policy updates.
\newblock In {\em 2017 IEEE international conference on robotics and automation
  (ICRA)}, pages 3389--3396. IEEE.

\bibitem[Gy{\"o}rfi et~al., 2006]{gyorfi2006distribution}
Gy{\"o}rfi, L., Kohler, M., Krzyzak, A., and Walk, H. (2006).
\newblock {\em A distribution-free theory of nonparametric regression}.
\newblock Springer Science \& Business Media.

\bibitem[Hajian and Domingo-Ferrer, 2012]{hajian2012methodology}
Hajian, S. and Domingo-Ferrer, J. (2012).
\newblock A methodology for direct and indirect discrimination prevention in
  data mining.
\newblock {\em IEEE transactions on knowledge and data engineering},
  25(7):1445--1459.

\bibitem[Hardt et~al., 2016]{hardt2016equality}
Hardt, M., Price, E., and Srebro, N. (2016).
\newblock Equality of opportunity in supervised learning.
\newblock {\em Advances in neural information processing systems}, 29.

\bibitem[H{\'e}bert-Johnson et~al., 2018]{hebert2018multicalibration}
H{\'e}bert-Johnson, U., Kim, M., Reingold, O., and Rothblum, G. (2018).
\newblock Multicalibration: Calibration for the (computationally-identifiable)
  masses.
\newblock In {\em International Conference on Machine Learning}, pages
  1939--1948. PMLR.

\bibitem[Heidari and Krause, 2018]{heidari2018preventing}
Heidari, H. and Krause, A. (2018).
\newblock Preventing disparate treatment in sequential decision making.
\newblock In {\em IJCAI}, pages 2248--2254.

\bibitem[Hort et~al., 2024]{hort2024bias}
Hort, M., Chen, Z., Zhang, J.~M., Harman, M., and Sarro, F. (2024).
\newblock Bias mitigation for machine learning classifiers: A comprehensive
  survey.
\newblock {\em ACM Journal on Responsible Computing}, 1(2):1--52.

\bibitem[Huber, 1981]{huber1981robust}
Huber, P.~J. (1981).
\newblock Robust statistics.
\newblock {\em Wiley Series in Probability and Mathematical Statistics}.

\bibitem[Janaro, 2023]{janaro2023nyc}
Janaro, C. (2023).
\newblock Nyc local law 144: A failed attempt at regulating ai in hiring.

\bibitem[Ji et~al., 2024]{ji2024beavertails}
Ji, J., Liu, M., Dai, J., Pan, X., Zhang, C., Bian, C., Chen, B., Sun, R.,
  Wang, Y., and Yang, Y. (2024).
\newblock Beavertails: Towards improved safety alignment of llm via a
  human-preference dataset.
\newblock {\em Advances in Neural Information Processing Systems}, 36.

\bibitem[Jiang and Nachum, 2020]{jiang2020identifying}
Jiang, H. and Nachum, O. (2020).
\newblock Identifying and correcting label bias in machine learning.
\newblock In {\em International conference on artificial intelligence and
  statistics}, pages 702--712. PMLR.

\bibitem[Kamiran and Calders, 2012]{kamiran2012data}
Kamiran, F. and Calders, T. (2012).
\newblock Data preprocessing techniques for classification without
  discrimination.
\newblock {\em Knowledge and information systems}, 33(1):1--33.

\bibitem[Kamishima et~al., 2012]{kamishima2012fairness}
Kamishima, T., Akaho, S., Asoh, H., and Sakuma, J. (2012).
\newblock Fairness-aware classifier with prejudice remover regularizer.
\newblock In {\em Machine Learning and Knowledge Discovery in Databases:
  European Conference, ECML PKDD 2012, Bristol, UK, September 24-28, 2012.
  Proceedings, Part II 23}, pages 35--50. Springer.

\bibitem[Kidder, 2001]{kidder2001does}
Kidder, W.~C. (2001).
\newblock Does the lsat mirror or magnify racial and ethnic differences in
  educational attainment: A study of equally achieving elite college students.
\newblock {\em Calif. L. Rev.}, 89:1055.

\bibitem[Kim et~al., 2018]{kim2018fairness}
Kim, M., Reingold, O., and Rothblum, G. (2018).
\newblock Fairness through computationally-bounded awareness.
\newblock {\em Advances in neural information processing systems}, 31.

\bibitem[Kleinberg et~al., 2016]{kleinberg2016inherent}
Kleinberg, J., Mullainathan, S., and Raghavan, M. (2016).
\newblock Inherent trade-offs in the fair determination of risk scores.
\newblock {\em arXiv preprint arXiv:1609.05807}.

\bibitem[Kleine~Buening et~al., 2022]{buening2021meritocracy}
Kleine~Buening, T., Segal, M., Basu, D., George, A.-M., and Dimitrakakis, C.
  (2022).
\newblock On meritocracy in optimal set selection.
\newblock In {\em Equity and Access in Algorithms, Mechanisms, and
  Optimization}, pages 1--14. ACM.

\bibitem[Lattimore and Szepesv{\'a}ri, 2020]{lattimore2020bandit}
Lattimore, T. and Szepesv{\'a}ri, C. (2020).
\newblock {\em \href{https://tor-lattimore.com/downloads/book/book.pdf}{Bandit
  algorithms}}.
\newblock Cambridge University Press.

\bibitem[Le~Merrer et~al., 2023]{le2023algorithmic}
Le~Merrer, E., Pons, R., and Tr{\'e}dan, G. (2023).
\newblock Algorithmic audits of algorithms, and the law.
\newblock {\em AI and Ethics}, pages 1--11.

\bibitem[Liu et~al., 2021]{liu2021adversarial}
Liu, J., Nogueira, M., Fernandes, J., and Kantarci, B. (2021).
\newblock Adversarial machine learning: A multilayer review of the
  state-of-the-art and challenges for wireless and mobile systems.
\newblock {\em IEEE Communications Surveys \& Tutorials}, 24(1):123--159.

\bibitem[Liu et~al., 2019]{liu2019implicit}
Liu, L.~T., Simchowitz, M., and Hardt, M. (2019).
\newblock The implicit fairness criterion of unconstrained learning.
\newblock In {\em International Conference on Machine Learning}, pages
  4051--4060. PMLR.

\bibitem[Liu et~al., 2022]{liu2022sample}
Liu, Q., Szepesv{\'a}ri, C., and Jin, C. (2022).
\newblock Sample-efficient reinforcement learning of partially observable
  markov games.
\newblock {\em Advances in Neural Information Processing Systems},
  35:18296--18308.

\bibitem[Liu et~al., 2017]{liu2017calibratedfairnessbandits}
Liu, Y., Radanovic, G., Dimitrakakis, C., Mandal, D., and Parkes, D.~C. (2017).
\newblock Calibrated fairness in bandits.

\bibitem[Lohia et~al., 2019]{lohia2019bias}
Lohia, P.~K., Ramamurthy, K.~N., Bhide, M., Saha, D., Varshney, K.~R., and
  Puri, R. (2019).
\newblock Bias mitigation post-processing for individual and group fairness.
\newblock In {\em Icassp 2019-2019 ieee international conference on acoustics,
  speech and signal processing (icassp)}, pages 2847--2851. IEEE.

\bibitem[Luong et~al., 2011]{luong2011k}
Luong, B.~T., Ruggieri, S., and Turini, F. (2011).
\newblock k-nn as an implementation of situation testing for discrimination
  discovery and prevention.
\newblock In {\em Proceedings of the 17th ACM SIGKDD international conference
  on Knowledge discovery and data mining}, pages 502--510.

\bibitem[Madiega, 2021]{madiega2021artificial}
Madiega, T. (2021).
\newblock Artificial intelligence act.
\newblock {\em European Parliament: European Parliamentary Research Service}.

\bibitem[Maneriker et~al., 2023]{maneriker2023online}
Maneriker, P., Burley, C., and Parthasarathy, S. (2023).
\newblock Online fairness auditing through iterative refinement.
\newblock In {\em Proceedings of the 29th ACM SIGKDD Conference on Knowledge
  Discovery and Data Mining}, KDD '23, page 1665–1676, New York, NY, USA.
  Association for Computing Machinery.

\bibitem[Mangold et~al., 2023]{mangold2023differential}
Mangold, P., Perrot, M., Bellet, A., and Tommasi, M. (2023).
\newblock Differential privacy has bounded impact on fairness in
  classification.
\newblock In {\em International Conference on Machine Learning}, pages
  23681--23705. PMLR.

\bibitem[Mehrabi et~al., 2021]{mehrabi2021survey}
Mehrabi, N., Morstatter, F., Saxena, N., Lerman, K., and Galstyan, A. (2021).
\newblock A survey on bias and fairness in machine learning.
\newblock {\em ACM computing surveys (CSUR)}, 54(6):1--35.

\bibitem[Mohri, 2018]{mohri2018foundations}
Mohri, M. (2018).
\newblock Foundations of machine learning.

\bibitem[Mutreja and Shafer, 2023]{mutreja2023pac}
Mutreja, S. and Shafer, J. (2023).
\newblock Pac verification of statistical algorithms.
\newblock In {\em The Thirty Sixth Annual Conference on Learning Theory}, pages
  5021--5043. PMLR.

\bibitem[Novelli et~al., 2023]{novelli2023taking}
Novelli, C., Casolari, F., Rotolo, A., Taddeo, M., and Floridi, L. (2023).
\newblock Taking ai risks seriously: a new assessment model for the ai act.
\newblock {\em AI \& SOCIETY}, pages 1--5.

\bibitem[Ouyang et~al., 2022]{ouyang2022training}
Ouyang, L., Wu, J., Jiang, X., Almeida, D., Wainwright, C., Mishkin, P., Zhang,
  C., Agarwal, S., Slama, K., Ray, A., et~al. (2022).
\newblock Training language models to follow instructions with human feedback.
\newblock {\em Advances in neural information processing systems},
  35:27730--27744.

\bibitem[Paltrinieri et~al., 2019]{paltrinieri2019learning}
Paltrinieri, N., Comfort, L., and Reniers, G. (2019).
\newblock Learning about risk: Machine learning for risk assessment.
\newblock {\em Safety science}, 118:475--486.

\bibitem[Pardau, 2018]{ccpa}
Pardau, S.~L. (2018).
\newblock The california consumer privacy act: Towards a european-style privacy
  regime in the united states.
\newblock {\em J. Tech. L. \& Pol'y}, 23:68.

\bibitem[Pentyala et~al., 2022]{pentyala2022privfairfl}
Pentyala, S., Neophytou, N., Nascimento, A., De~Cock, M., and Farnadi, G.
  (2022).
\newblock Privfairfl: Privacy-preserving group fairness in federated learning.
\newblock {\em arXiv preprint arXiv:2205.11584}.

\bibitem[Pessach and Shmueli, 2022]{pessach2022review}
Pessach, D. and Shmueli, E. (2022).
\newblock A review on fairness in machine learning.
\newblock {\em ACM Computing Surveys (CSUR)}, 55(3):1--44.

\bibitem[Pleiss et~al., 2017]{pleiss2017fairness}
Pleiss, G., Raghavan, M., Wu, F., Kleinberg, J., and Weinberger, K.~Q. (2017).
\newblock On fairness and calibration.
\newblock {\em Advances in neural information processing systems}, 30.

\bibitem[Salimi et~al., 2019a]{salimi2019data}
Salimi, B., Howe, B., and Suciu, D. (2019a).
\newblock Data management for causal algorithmic fairness.
\newblock {\em arXiv preprint arXiv:1908.07924}.

\bibitem[Salimi et~al., 2019b]{salimi2019capuchin}
Salimi, B., Rodriguez, L., Howe, B., and Suciu, D. (2019b).
\newblock Capuchin: Causal database repair for algorithmic fairness.
\newblock {\em arXiv preprint arXiv:1902.08283}.

\bibitem[Salimi et~al., 2019c]{salimi2019interventional}
Salimi, B., Rodriguez, L., Howe, B., and Suciu, D. (2019c).
\newblock Interventional fairness: Causal database repair for algorithmic
  fairness.
\newblock In {\em Proceedings of the 2019 International Conference on
  Management of Data}, pages 793--810.

\bibitem[Shankar et~al., 2024]{shankar2024validates}
Shankar, S., Zamfirescu-Pereira, J., Hartmann, B., Parameswaran, A., and
  Arawjo, I. (2024).
\newblock Who validates the validators? aligning llm-assisted evaluation of llm
  outputs with human preferences.
\newblock In {\em Proceedings of the 37th Annual ACM Symposium on User
  Interface Software and Technology}, pages 1--14.

\bibitem[Shapley, 1953]{shapley1953stochastic}
Shapley, L.~S. (1953).
\newblock Stochastic games.
\newblock {\em Proceedings of the national academy of sciences},
  39(10):1095--1100.

\bibitem[Shukla and Basu, 2024]{shukla:hal-04733134}
Shukla, A. and Basu, D. (2024).
\newblock {Preference-based Pure Exploration}.
\newblock In {\em {Advances in Neural Information Processing Systems
  (NeurIPS)}}, Vancouver (CA), Canada.

\bibitem[Silver et~al., 2016]{silver2016mastering}
Silver, D., Huang, A., Maddison, C.~J., Guez, A., Sifre, L., Van Den~Driessche,
  G., Schrittwieser, J., Antonoglou, I., Panneershelvam, V., Lanctot, M.,
  et~al. (2016).
\newblock Mastering the game of go with deep neural networks and tree search.
\newblock {\em nature}, 529(7587):484--489.

\bibitem[Silver et~al., 2017]{silver2017mastering}
Silver, D., Hubert, T., Schrittwieser, J., Antonoglou, I., Lai, M., Guez, A.,
  Lanctot, M., Sifre, L., Kumaran, D., Graepel, T., et~al. (2017).
\newblock Mastering chess and shogi by self-play with a general reinforcement
  learning algorithm.
\newblock {\em arXiv preprint arXiv:1712.01815}.

\bibitem[Song et~al., 2024]{song2024preference}
Song, F., Yu, B., Li, M., Yu, H., Huang, F., Li, Y., and Wang, H. (2024).
\newblock Preference ranking optimization for human alignment.
\newblock In {\em Proceedings of the AAAI Conference on Artificial
  Intelligence}, volume~38, pages 18990--18998.

\bibitem[Sorin, 1986]{sorin1986asymptotic}
Sorin, S. (1986).
\newblock Asymptotic properties of a non-zero sum stochastic game.
\newblock {\em International Journal of Game Theory}, 15:101--107.

\bibitem[Sun et~al., 2021]{sun2021probabilistic}
Sun, B., Sun, J., Dai, T., and Zhang, L. (2021).
\newblock Probabilistic verification of neural networks against group fairness.

\bibitem[Sutton and Barto, 2018]{sutton2018reinforcement}
Sutton, R.~S. and Barto, A.~G. (2018).
\newblock {\em
  \href{https://www.andrew.cmu.edu/user/rmorina/papers/SuttonBook.pdf}{Reinforcement
  learning: An introduction}}.
\newblock MIT press.

\bibitem[Tao et~al., 2024]{cultural_bias}
Tao, Y., Viberg, O., Baker, R.~S., and Kizilcec, R.~F. (2024).
\newblock {Cultural bias and cultural alignment of large language models}.
\newblock {\em PNAS Nexus}, 3(9):pgae346.

\bibitem[Tavara et~al., 2021]{dpadmm}
Tavara, S., Schliep, A., and Basu, D. (2021).
\newblock
  \href{https://drive.google.com/file/d/1ejIJsPboED_jDSI59RITSwXW78ZmzyX1/view}{Federated
  learning of oligonucleotide drug molecule thermodynamics
  with differentially private {ADMM}-based {SVM}}.
\newblock In {\em Machine Learning and Principles and Practice of Knowledge
  Discovery in Databases}, pages 459--467.

\bibitem[Vapnik, 1991]{vapnik1991principles}
Vapnik, V. (1991).
\newblock Principles of risk minimization for learning theory.
\newblock {\em Advances in neural information processing systems}, 4.

\bibitem[Veale and Zuiderveen~Borgesius, 2021]{veale2021demystifying}
Veale, M. and Zuiderveen~Borgesius, F. (2021).
\newblock Demystifying the draft eu artificial intelligence act—analysing the
  good, the bad, and the unclear elements of the proposed approach.
\newblock {\em Computer Law Review International}, 22(4):97--112.

\bibitem[Voigt and Von~dem Bussche, 2017]{gdpr}
Voigt, P. and Von~dem Bussche, A. (2017).
\newblock The eu general data protection regulation (gdpr).
\newblock {\em A Practical Guide, 1st Ed., Cham: Springer International
  Publishing}, 10(3152676):10--5555.

\bibitem[Wang et~al., 2020]{wang2020dp}
Wang, B., Gu, Q., Boedihardjo, M., Wang, L., Barekat, F., and Osher, S.~J.
  (2020).
\newblock Dp-lssgd: A stochastic optimization method to lift the utility in
  privacy-preserving erm.
\newblock In {\em Mathematical and Scientific Machine Learning}, pages
  328--351. PMLR.

\bibitem[Wang et~al., 2023]{wang2023aligning}
Wang, Y., Zhong, W., Li, L., Mi, F., Zeng, X., Huang, W., Shang, L., Jiang, X.,
  and Liu, Q. (2023).
\newblock Aligning large language models with human: A survey.
\newblock {\em arXiv preprint arXiv:2307.12966}.

\bibitem[Wright et~al., 2024]{wright2024null}
Wright, L., Muenster, R.~M., Vecchione, B., Qu, T., Cai, P., Smith, A.,
  Investigators, C. .~S., Metcalf, J., Matias, J.~N., et~al. (2024).
\newblock Null compliance: Nyc local law 144 and the challenges of algorithm
  accountability.
\newblock In {\em The 2024 ACM Conference on Fairness, Accountability, and
  Transparency}, pages 1701--1713.

\bibitem[Xiao et~al., 2024]{Preference_collapse}
Xiao, J., Li, Z., Xie, X., Getzen, E., Fang, C., Long, Q., and Su, W.~J.
  (2024).
\newblock On the algorithmic bias of aligning large language models with rlhf:
  Preference collapse and matching regularization.
\newblock {\em arXiv preprint arXiv:2405.16455}.

\bibitem[Yan and Zhang, 2022]{yan2022active}
Yan, T. and Zhang, C. (2022).
\newblock Active fairness auditing.
\newblock In {\em International Conference on Machine Learning}, pages
  24929--24962. PMLR.

\bibitem[Yang et~al., 2020]{yang2020multi}
Yang, Y., Juntao, L., and Lingling, P. (2020).
\newblock Multi-robot path planning based on a deep reinforcement learning dqn
  algorithm.
\newblock {\em CAAI Transactions on Intelligence Technology}, 5(3):177--183.

\bibitem[Yu et~al., 2024]{yu2024rlhf}
Yu, T., Yao, Y., Zhang, H., He, T., Han, Y., Cui, G., Hu, J., Liu, Z., Zheng,
  H.-T., Sun, M., et~al. (2024).
\newblock Rlhf-v: Towards trustworthy mllms via behavior alignment from
  fine-grained correctional human feedback.
\newblock In {\em Proceedings of the IEEE/CVF Conference on Computer Vision and
  Pattern Recognition}, pages 13807--13816.

\bibitem[Zemel et~al., 2013]{zemel2013learning}
Zemel, R., Wu, Y., Swersky, K., Pitassi, T., and Dwork, C. (2013).
\newblock Learning fair representations.
\newblock In {\em International conference on machine learning}, pages
  325--333. PMLR.

\bibitem[Zhang et~al., 2020]{zhang2020model}
Zhang, K., Kakade, S., Basar, T., and Yang, L. (2020).
\newblock Model-based multi-agent rl in zero-sum markov games with near-optimal
  sample complexity.
\newblock {\em Advances in Neural Information Processing Systems},
  33:1166--1178.

\bibitem[Zhang et~al., 2024]{Genderalign}
Zhang, T., Zeng, Z., Xiao, Y., Zhuang, H., Chen, C., Foulds, J., and Pan, S.
  (2024).
\newblock Genderalign: An alignment dataset for mitigating gender bias in large
  language models.
\newblock {\em arXiv preprint arXiv:2406.13925}.

\bibitem[Ziegler et~al., 2019]{ziegler2019fine}
Ziegler, D.~M., Stiennon, N., Wu, J., Brown, T.~B., Radford, A., Amodei, D.,
  Christiano, P., and Irving, G. (2019).
\newblock Fine-tuning language models from human preferences.
\newblock {\em arXiv preprint arXiv:1909.08593}.

\end{thebibliography}

\end{document}